%% file: root.tex
\documentclass[letterpaper, 10 pt, conference]{ieeeconf}  

\IEEEoverridecommandlockouts                              

\usepackage{hyperref}
\usepackage{url}
\usepackage{booktabs}
\usepackage{amsfonts}
\usepackage{nicefrac}
\usepackage{microtype}
\usepackage{xcolor}
\usepackage{multirow}
\usepackage{graphicx}
\usepackage{adjustbox}
\usepackage{amsmath}
\usepackage{amssymb}
\usepackage{colortbl}
\usepackage{makecell}
\usepackage{caption}
\usepackage{subcaption}
\usepackage{wrapfig}
\usepackage{tikz}
\usepackage{booktabs}
\usepackage{lipsum}
\let\labelindent\relax
\usepackage{enumitem}
\usepackage{bm}

\definecolor{normalblue}{RGB}{31,119,180}
\definecolor{hallorange}{RGB}{255,127,14}
\usetikzlibrary{arrows.meta}

\title{\LARGE \bf
Sparse Feature Policy Unlearning\\Mitigates State Hallucination in Vision-Language-Action Models
}

\author{Jiho Lee$^{1}$, Jeongeun Park$^{2}$, Heayoun Choi$^{1}$, Taekyung Kim$^{3,*}$, and Eunwoo Kim$^{1,*}$
\thanks{$^{1}$Jiho Lee, Heayoun Choi, and Eunwoo Kim are with the School of Computer Science Engineering, Chung-Ang University, Seoul, 06974, Republic of Korea (e-mail: \{j2hoooo, heayounchoi, eunwoo\}@cau.ac.kr).}
\thanks{$^{2}$Jeongeun Park is with the School of Interactive Computing, Georgia Institute of Technology, Atlanta, GA, 30308, USA (e-mail: jpark3456@gatech.edu)}
\thanks{$^{3}$Taekyung Kim is with the NAVER AI Lab, Seongnam, 13561, Republic of Korea (e-mail: taekyung.k@navercorp.com). $^{*}$Corresponding authors.}
}

\begin{document}

\maketitle
\thispagestyle{empty}
\pagestyle{empty}

\begin{abstract}
Vision-Language-Action (VLA) models have shown strong generalization in robotic manipulation by leveraging rich representations from pretrained vision-language models.
However, their deployment in real-world environments remains limited by recurring unreliable behaviors.
In this work, we study \textit{state hallucination}, a recurring failure pattern in which a VLA continues acting as if an unrealized robot-object state had been achieved.
Our analyses find that state hallucination coincides with weakened attention to task-relevant visual regions, and a mechanistic interpretation via sparse autoencoders reveals that hallucination-associated sparse features are activated when these failures occur.
Based on this analysis, we propose \textit{SOUL} (Sparse feature pOlicy UnLearning), which selectively unlearns policy knowledge associated with state hallucination behaviors, where sparse features identified from hallucination failures and successful behaviors serve as explicit forgetting and retention targets, respectively.
Experiments across VLA architectures in simulated and real-world environments show that our method substantially reduces hallucinated failures and improves task success without substantially compromising the existing manipulation capabilities.
These results suggest that interpretable feature analysis provides a practical basis for selectively modifying undesirable knowledge in robot policies.
\end{abstract}

\section{INTRODUCTION}
Vision-language-action models (VLAs) have emerged as a promising approach for bringing foundation models into robot manipulation~\cite{zitkovich2023rt2,kim2025openvla,intelligence2025pi_}.
By integrating visual and language understanding with action generation, VLAs can translate high-level instructions into continuous robot control.
Large-scale pretraining on heterogeneous robot and vision-language data has further enabled generalization across tasks, objects, environments, and robot embodiments~\cite{oneill2023openx,gr00tn1_2025,gemini2025robotics}.
Despite this progress, VLAs still exhibit unreliable behaviors, including unstable or premature actions, long-horizon drift, failures to recover from execution errors, and infeasible actions~\cite{gu2026safe,zhang2026safevla,fei2025liberoplus,zhou2025libero,zhang2025experiences,lin2025failsafe,soh2026actionhallucination}.
This is particularly concerning in physical robot execution, where erroneous actions directly affect the robot and its environment.

Within this broader set of failures, we focus on a recurring failure pattern in which a VLA continues a manipulation sequence based on a robot-object state that has not actually been achieved.
We refer to this behavioral pattern as \textbf{state hallucination}, because the policy acts as if an unrealized physical state were true.
For example, a VLA may close its gripper before reaching an object and subsequently execute transport and place motions as though the object has been successfully grasped, as shown in Fig.~\ref{fig:1_teaser}\subref{fig:1_teaser_ex}.
Such a mismatch can persist across subsequent actions, allowing a local execution error to propagate over multiple stages.
This recurring and persistent behavior motivates us to investigate whether policy knowledge associated with state hallucination can be selectively removed while preserving successful manipulation capabilities.

%
%
\input{Figures/Tex/1_teaser}
%
%

We first examine how the policy's internal representations differ between hallucinated and successful behaviors.
Such hallucinated rollouts exhibit weakened attention to task-relevant regions, such as the gripper and target object, together with more spatially dispersed attention.
Through a mechanistic analysis with sparse autoencoders, we also observe that sparse features exhibit distinct activation patterns during hallucinated and normal executions.
To characterize these sparse feature activation patterns, we define a feature association score that contrasts activations across execution outcomes, such as failures with hallucination, failures without hallucination, and successful executions.
Based on this score, we categorize sparse features according to whether they are more strongly associated with hallucinated failures or successful executions.

Building on these observations, we propose \emph{\textbf{S}parse feature p\textbf{O}licy \textbf{U}n\textbf{L}earning} (\textbf{SOUL}), which selectively unlearns policy knowledge associated with state hallucination, where sparse features identified from hallucinated failures and successful behaviors serve as explicit forgetting and retention targets, respectively.
This feature-level policy unlearning objective suppresses activations of the forgetting targets while enhancing activations of the retention targets, thereby mitigating recurring hallucination behavior while minimizing interference with useful manipulation capabilities.
We validate our approach with OpenVLA~\cite{kim2025openvla} and $\pi_{0.5}$~\cite{intelligence2025pi_} on LIBERO-Plus~\cite{fei2025liberoplus}, RoboCasa~\cite{nasiriany2024robocasa}, and a real-world Franka robot, achieving an average {21.7}\%p reduction in hallucination rate and a {22.1}\%p improvement in task success while preserving 89\% of behaviors that were successful before unlearning.
These results demonstrate that interpretable feature analysis can reveal internal representations associated with recurring policy failures, providing a practical basis for selectively unlearning such behaviors in embodied foundation models.
Our contributions are as follows:
\begin{itemize}[leftmargin=*]
    \item We characterize state hallucination as a recurring failure pattern in which a VLA continues acting as though an unrealized robot-object state has been achieved.
    \item We analyze state hallucination through visual attention and sparse features, revealing weakened attention to task-relevant regions and distinct feature patterns associated with state hallucination.
    \item We introduce a sparse feature policy unlearning method that selectively unlearns recurring state hallucination behavior by suppressing hallucination features while enhancing features that support successful manipulation.
\end{itemize}

\section{Related Work}
\subsection{Vision-Language-Action Models}
Vision-language-action models have become a major direction for generalist robot manipulation. 
Prior work has improved generalization through larger pretrained vision-language representations, heterogeneous robotic data, and stronger action generation architectures~\cite{zitkovich2023rt2,kim2025openvla,intelligence2025pi_,oneill2023openx,gr00tn1_2025,gemini2025robotics,driess2023palme,octo2024,black2024pi0,shukor2025smolvla}. 
Deployment reliability has also been improved through intermediate reasoning~\cite{zhao2025cotvla,lin2026onetwovla}, failure detection and recovery~\cite{lin2025failsafe,liu2026selfcorrectingvla}, and test-time sampling or verification~\cite{kwok2025robomonkey,jang2025mgselect}.
These approaches primarily focus on improving policy outputs during deployment, rather than directly modifying policy knowledge associated with recurring failures.
Recent concurrent mechanistic interpretability studies have begun to analyze and steer VLA behavior at inference time by intervening on representations, such as feed-forward activation directions~\cite{haon2025mechanistic} and sparse features~\cite{khan2025controlling}.

In contrast, we formulate the selective modification of undesirable policy behavior as a policy unlearning problem, where knowledge associated with recurring hallucinated behavior is removed while representations supporting successful execution are preserved.
Encoding this modification in the pretrained policy parameters enables persistent behavior correction without repeated intervention at inference time.

\subsection{Machine Unlearning}
Machine unlearning aims to make a trained model forget knowledge associated with selected data without retraining the model from scratch~\cite{cao2015unlearning,nguyen2025survey}.
Prior work has studied unlearning mainly in classification tasks, where the influence of selected training data is removed from trained models~\cite{golatkar2020eternal,thudi2022unrolling}, and in language tasks, where factual knowledge or undesired responses are suppressed~\cite{yao2024large,maini2024tofu}.
A central challenge is balancing the removal of targeted knowledge with the preservation of useful model capabilities. 
This trade-off is particularly consequential for robot policies, where modifying one behavior can inadvertently disrupt multiple tightly coupled skills.
Motivated by this, we introduce feature-level selective policy unlearning by explicitly identifying features for forgetting and retention to suppress undesirable behavior while preserving useful manipulation capabilities.

\subsection{Sparse Autoencoders}
Sparse autoencoders (SAEs) decompose dense model activations into sparse latent representations, enabling individual features to be associated with more localized and interpretable patterns in model behavior~\cite{huben2024sparse,gao2024scaling,bussmann2024batchtopk}.
They are trained to reconstruct the original activations under a sparsity constraint, such that only a small subset of latent features is active for each input. 
A commonly used variant is the Top-$k$ SAE, which imposes this sparsity constraint explicitly by retaining only the $k$ largest latent activations.
In VLAs, SAE features have recently been used to interpret and repeatedly steer policy at test time~\cite{khan2025controlling}.
In contrast, the proposed method identifies recurring internal patterns associated with unreliable behaviors using sparse features and translates these patterns into forgetting targets to guide targeted policy unlearning.

%
%
\input{Figures/Tex/2_attention}
%
%

\section{Analysis of State Hallucination in VLAs}
\label{sec:analysis}
\subsection{What Is State Hallucination?}
We define \textbf{state hallucination} as a behavioral execution pattern in which a VLA continues to generate actions based on a robot-object state that has not actually been achieved.
State hallucination can occur at different manipulation stages:
\begin{itemize}[leftmargin=*]
    \item A \emph{grasp hallucination} occurs when the gripper closes in free space without contacting the target object.
    Note that grasp failures involving object contact, such as inaccurate localization or slippage, are not considered hallucinations.
    \item A \emph{transport hallucination} occurs when, after either a grasp failure or a grasp hallucination, the gripper remains closed and continues moving from the grasp location toward the destination while the object remains ungrasped.
    \item A \emph{place hallucination} occurs when the gripper opens at the destination without holding the target object.
\end{itemize}
We label these events by tracking ground-truth robot-object states in simulation and observable interaction states in the real world.
Based on these criteria, we categorize each rollout as clean success (CS), hallucination failure (HF), or non-hallucination failure (NF).
Here, HF denotes failed executions that exhibit at least one state hallucination event, whereas CS and NF denote successful and failed executions without state hallucination, respectively.

\subsection{Attention Analysis of Hallucinated and Normal Executions}
\label{sec:attn_anal}
To examine how the policy behaves internally during state hallucination, we compare internal attention patterns between CS and HF executions.
At each rollout step, we aggregate the attention from action tokens to visual tokens and project the resulting attention scores onto the input image, revealing which visual regions the policy attends to while generating actions.
For visualization, we show representative time steps immediately preceding the corresponding manipulation event, such as grasping.
As shown in Fig.~\ref{fig:2_attention}, CS executions exhibit attention concentrated on task-relevant regions, particularly the target object and gripper.
In contrast, HF executions show weaker attention to these regions and increased attention to background regions.
These differences suggest that state hallucination is accompanied by a distinct shift in the policy's visual attention. 
This motivates us to look beyond attention and examine whether hallucinated executions are also reflected in the internal representations of the policy.

%
%
\input{Figures/Tex/3_feature}
%
%

\subsection{Sparse Feature Characterization of State Hallucination}
\label{sec:sae_anal}
To investigate whether state hallucination is reflected in distinguishable internal representations of VLAs, we analyze policy activations using sparse autoencoders (SAEs)~\cite{huben2024sparse}.
SAEs decompose dense representations into sparse features, allowing us to characterize how individual feature activations vary across different behaviors, such as clean success, hallucination failure, and non-hallucination failure.
Through this characterization, we find that sparse features are selectively associated with state hallucination and successful behaviors.

\textbf{SAE training.}
To extract sparse features from policy activations, we collect visual-token activations from the residual streams of transformer layers throughout each rollout together with their execution outcome labels.
The visual token activations allow sparse features to be examined with respect to semantic scene regions.
For an input sample $x$, let $x_p\in\mathbb{R}^{d}$ denote the activation corresponding to visual token $p$.
We flatten the collected activations across rollout steps and visual tokens, treating each $x_p$ as an individual SAE training sample.
The encoder maps $x_p$ to a latent representation and retains only the $k$ largest positive activations, yielding a sparse feature vector $z_p(x)$, whose $j$-th element $z_{p,j}(x)$ denotes the activation of SAE feature $j$ at visual token $p$.
The decoder reconstructs the original policy activation $x_p$ from $z_p(x)$.
After training, we freeze the SAE and use its encoder to extract sparse feature activations for subsequent analysis.

\textbf{Sparse feature categorization.}
Motivated by the region-level attention differences observed in Section~\ref{sec:attn_anal}, we characterize how sparse features are activated across semantic scene regions.
We obtain semantic regions from the masks provided by the simulation benchmarks and manually annotate the corresponding regions for real-world observations.
For an input sample $x$, let $\mathcal{R}_r(x)$ denote the set of visual tokens belonging to region $r$.
We define the region-conditioned feature activation as:
\begin{equation}
    z_{r,j}(x)
    =
    \frac{1}{|\mathcal{R}_r(x)|}
    \sum_{p \in \mathcal{R}_r(x)}
    z_{p,j}(x).
\end{equation}
This quantity measures how strongly feature $j$ is expressed within semantic region $r$.
For rollout $e$ at time step $t$, we denote the corresponding region-conditioned feature activation by $z_{e,t,r,j}$.
Since individual execution events, such as grasp, may occur over limited portions of a rollout, we summarize the temporal activation of each region-feature pair by its maximum over time, $m_{e,r,j}=\max\limits_t z_{e,t,r,j}.$
We then average these rollout-level activations within each execution outcome.
Let $\mathcal{E}_c$ denote the set of rollouts belonging to outcome
$c \in \{\mathrm{CS},\mathrm{NF},\mathrm{HF}\}$.
The outcome-level activation is computed as:
\begin{equation}
    \bar{m}_{c,r,j}
    =
    \frac{1}{|\mathcal{E}_c|}
    \sum_{e \in \mathcal{E}_c}
    m_{e,r,j}.
\end{equation}
To examine how individual sparse features are associated with different execution outcomes, we define the following feature association scores:
\begin{equation}
\begin{aligned}
    s^{\mathrm{hall}}_{r,j}
    &=
    \bar{m}_{\mathrm{HF},r,j}
    -
    \frac{1}{2}
    \left(
    \bar{m}_{\mathrm{CS},r,j}
    +
    \bar{m}_{\mathrm{NF},r,j}
    \right), \\
    {s^{\mathrm{succ}}_{r,j}}
    &=
    \bar{m}_{\mathrm{CS},r,j}
    -
    \frac{1}{2}
    \left(
    \bar{m}_{\mathrm{HF},r,j}
    +
    \bar{m}_{\mathrm{NF},r,j}
    \right).
\end{aligned}
\end{equation}
Here, larger \(s^{\mathrm{hall}}_{r,j}\) and \(s^{\mathrm{succ}}_{r,j}\) indicate that feature \(j\) at region \(r\) is more strongly activated in hallucination failures and clean successes, respectively, relative to the other outcomes.
Based on these scores, we categorize sparse features according to their stronger association with hallucination failures or clean successes, and refer to them as hallucination features and success features, respectively.

\textbf{Sparse features are selectively associated with state hallucination.}
Fig.~\ref{fig:3_feature} visualizes representative hallucination and success features across HF and CS rollouts, showing both their temporal activations and the distributions of their rollout-level activations.
For each rollout, the temporal activation \(z_{e,t,r,j}\) shows how the sparse feature activation changes over time throughout the rollout.
The hallucination feature remains weak during CS rollouts, but becomes strongly activated around state hallucination events in HF rollouts.
Its activation is concentrated around hallucinated execution periods rather than remaining uniformly high throughout the rollout, indicating that the feature captures a temporally localized pattern associated with state hallucination.
Across HF rollouts spanning different tasks and visual scenes, similar activation patterns repeatedly appear around hallucination events, revealing a consistent feature pattern associated with the behavior.
Conversely, the success feature becomes strongly activated during successful subtask execution or task completion in CS rollouts, while remaining weaker in HF.

Beyond these trajectory-level representative rollouts, boxplots on the right summarize the distributions of the rollout-level maximum activation \(m_{e,r,j}\) across CS and HF rollouts.
Consistent with the temporal activation patterns, the hallucination feature shows higher activation in HF, whereas the success feature shows higher activation in CS.
These differences remain consistent across tasks and visual conditions.
Together, these results show that state hallucination and successful execution are associated with distinguishable sparse feature patterns within the VLA representation.
More visualizations are provided in Section~\ref{sec:exp} and the supplementary materials.

%
%
\input{Figures/Tex/4_overview}
%
%

\section{\textbf{S}parse {F}eature P\textbf{o}licy \textbf{U}n\textbf{l}earning (\textbf{SOUL})}
\label{sec:soul}
Building on the sparse feature analysis, we formulate the state hallucination mitigation as a selective policy unlearning problem.
Specifically, internal representations associated with state hallucination are suppressed while those supporting successful execution are enhanced.
To this end, we target sparse features associated with hallucination failure and clean success as explicit forgetting and retention, respectively.
The overall framework is illustrated in Fig.~\ref{fig:fig_4_overview}.

\textbf{Feature selection for forgetting and retention.}
We construct the forget and retain sets using the feature association scores defined in Section~\ref{sec:sae_anal}.
For forgetting, we prioritize features strongly associated with hallucination failures, distinguishable from non-hallucination failures, and weakly associated with clean success.
The first hallucination association is captured by \( s^{\mathrm{hall}}_{r,j} \), while the other two criteria are combined as \( s^{\mathrm{agg}}_{r,j}=s^{\mathrm{sep}}_{r,j}-s^{\mathrm{succ}}_{r,j} \), where \( s^{\mathrm{sep}}_{r,j}=\bar{m}_{\mathrm{HF},r,j}-\bar{m}_{\mathrm{NF},r,j} \) measures separation from non-hallucination failures and subtracting \( s^{\mathrm{succ}}_{r,j} \) penalizes association with successful execution.
We then define the final forget score as \( s^{\mathcal F}_{r,j}=s^{\mathrm{hall}}_{r,j}+s^{\mathrm{agg}}_{r,j} \) and select the top-ranked features to form the forget set \( \mathcal F \).
For retention, we select features with high $s^{\mathrm{succ}}_{r,j}$ as the retain set $\mathcal{R}$, favoring features strongly associated with clean success.
The resulting sets $\mathcal{F}$ and $\mathcal{R}$ define the sparse feature targets to be suppressed and enhanced, respectively, during policy unlearning.

\textbf{Feature-guided policy unlearning.}
Given the forget set $\mathcal{F}$ and retain set $\mathcal{R}$, we update the VLA policy using a feature-guided unlearning objective.
During unlearning, the SAE encoder remains frozen and is used only to compute sparse feature activations, while the VLA policy parameters $\theta$ are optimized.
For hallucination failure samples, we define the forgetting objective to penalize activations of the selected forget features as:
\begin{equation}
    \mathcal{L}_{\mathrm{forget}}
    =
    \mathbb{E}_{x \in \mathcal{D}_{\mathrm{HF}}}
    \left[
    \frac{1}{|\mathcal{F}|}
    \sum_{(r,j) \in \mathcal{F}}
    \left(z_{r,j}(x;\theta)\right)^2
    \right].
\end{equation}
For clean success samples, we define the retention objective to encourage activation of the selected retain features as:
\begin{equation}
    \mathcal{L}_{\mathrm{retain}}
    =
    -
    \mathbb{E}_{x \in \mathcal{D}_{\mathrm{CS}}}
    \left[
    \frac{1}{|\mathcal{R}|}
    \sum_{(r,j) \in \mathcal{R}}
    z_{r,j}(x;\theta)
    \right].
\end{equation}
In summary, the forgetting objective suppresses feature activations associated with state hallucination, while the retention objective reinforces features associated with successful execution, helping preserve useful policy representations during policy unlearning.
The final objective jointly balances forgetting and retention as
$\mathcal{L}_{\mathrm{unlearn}}
=
\mathcal{L}_{\mathrm{forget}}
+
\lambda_{\mathrm{r}}
\mathcal{L}_{\mathrm{retain}},$
where $\lambda_{\mathrm{r}}$ controls the strength of the retention objective.

%
%
\input{Tables/1_main}
%
%

%
%
\input{Figures/Tex/5_benchmark_rollouts}
%
%

\section{Experiments}
\label{sec:exp}
\subsection{Setup} 
\label{sec:exp_details}
\textbf{Evaluation protocol.}
We conduct experiments with the autoregressive OpenVLA~\cite{kim2025openvla} and flow-matching-based $\pi_{0.5}$~\cite{intelligence2025pi_} in LIBERO-Plus~\cite{fei2025liberoplus}, RoboCasa~\cite{nasiriany2024robocasa}, and a real-world setup with a 7-DoF Franka Research 3 robotic arm.
LIBERO-Plus provides realistic visual perturbations at multiple severity levels, while RoboCasa provides diverse kitchen environments and everyday manipulation tasks.
To provide a controlled setting for evaluating state hallucination while minimizing confounding effects from task and motion planning, we focus on pick-and-place tasks, a fundamental manipulation primitive underlying many complex tasks, with a structured grasp-transport-place sequence.\footnote{
For LIBERO-Plus, we use the base tasks \emph{put the bowl on the stove}, \emph{pick up the black bowl on the cookie box and place it on the plate}, and \emph{pick up the tomato sauce and place it in the basket} from the goal, spatial, and object suites, respectively, together with their corresponding lighting and background evaluation variants. 
For RoboCasa, we use \emph{PickPlaceCounterToStandMixer}, \emph{PickPlaceDrawerToCounter}, \emph{PickPlaceCounterToSink}, \emph{PickPlaceCabinetToCounter}, \emph{PickPlaceCounterToCabinet}, and \emph{PickPlaceFridgeDrawerToShelf}, together with their corresponding evaluation configurations.}
To our knowledge, no prior work has established an unlearning baseline for robot policies. 
For a direct comparison from an unlearning perspective, we therefore adapt gradient ascent, a commonly used optimization-based unlearning method~\cite{yao2024large,maini2024tofu}, as a baseline for VLA policies by maximizing the action prediction loss on hallucination failure samples.

\textbf{SAE-based feature analysis and unlearning.}
We use Top-$k$ SAEs~\cite{gao2024scaling,makhzani2013ksparse} to decompose visual-token activations into sparse latent features, whose fixed number of active features provides a consistent basis for comparing activation patterns across executions.
The SAE objective combines the standard reconstruction loss with an auxiliary reconstruction loss using the top-$k_{\mathrm{aux}}$ inactive features to provide additional learning signals for rarely activated features.
We set the SAE latent dimension to $4096$, the number of active features to $k=100$ and $k_{\mathrm{aux}}=512$.
From the sparse feature analysis, we select the top five region-feature pairs associated with hallucination and success as forgetting and retention targets, respectively.
For policy unlearning, we initialize the model from the publicly released pretrained policy and optimize LoRA adapters~\cite{hu2021lora} with our unlearning objective while keeping the original model weights frozen.
We uniformly sample up to $20$ time steps per rollout to reduce bias toward longer trajectories.
For feature analysis, we use layers 0, 8, 16, 24, and 31 of the OpenVLA language backbone and layers 0, 5, 11, and 17 of the PaliGemma backbone in $\pi_{0.5}$.

\textbf{Metrics.} 
To evaluate task performance and state hallucination, we report task-level success metrics and rollout-level hallucination metrics.
\textit{Overall success} and \textit{clean success} measure the fraction of rollouts that successfully complete the task and the fraction that complete the task without state hallucination, respectively.
\textit{Hallucination} and \textit{hallucination failure} measure the fraction of rollouts containing at least one state hallucination event and the fraction that both contain state hallucination and result in task failure, respectively.
We further report type-specific hallucination rates for \textit{grasp}, \textit{transport}, and \textit{place}, defined as the fraction of rollouts containing at least one hallucination of each type.

\subsection{Main Results}
\label{sec:exp_results}
{\textbf{Quantitative results.}}
To quantitatively evaluate whether our proposed sparse feature policy unlearning (SOUL) improves policy reliability by reducing state hallucination without compromising task success, we compare three policies: the original pretrained policy, gradient-ascent unlearning, and SOUL.
As shown in TABLE~\ref{tab:tab_1}, the original policies exhibit average hallucination and hallucination failure rates of $55.0\%$ and $48.6\%$ across evaluation settings, respectively.
Gradient-ascent unlearning does not consistently mitigate hallucination and substantially degrades task success, suggesting that directly forgetting hallucination samples can interfere with useful manipulation capabilities.
In contrast, SOUL reduces the average hallucination rate from $55.0\%$ to $33.5\%$, while increasing clean success from $27.5\%$ to $52.2\%$.
Hallucination rates of the grasp, transport, and place stages are also consistently reduced, showing that the improvement extends across manipulation stages rather than being limited to one stage.
These results show that SOUL improves policy reliability by selectively mitigating recurring state hallucination behaviors while preserving useful manipulation capabilities.

%
%
\input{Figures/Tex/6_real_rollouts}
%
%

{\textbf{Qualitative results.}}
To qualitatively examine how SOUL modifies sparse feature activations, Fig.~\ref{fig:fig_5} compares rollout behaviors and the temporal activations of selected forget and retain features before and after applying unlearning across the evaluation settings.
Before unlearning, the original policies exhibit state hallucination, with strong forget feature activation emerging around hallucination events while retain feature activation remains weak.
After unlearning, the policies successfully complete the tasks without state hallucination, accompanied by suppressed forget feature activation and increased retain feature activation during the corresponding manipulation stages.
This pattern is consistently observed for both OpenVLA on LIBERO-Plus and $\pi_{0.5}$ on RoboCasa.
These results show that SOUL shifts the selected sparse feature activations in the intended directions, suppressing the activation of hallucination-associated features while increasing that of features associated with clean successful execution.

{\textbf{Real-world evaluation.}}
We further evaluate SOUL on a Franka Research 3 robot using OpenVLA and $\pi_{0.5}$, conducting 50 rollouts across both models under diverse visual conditions.
Starting from the publicly released pretrained models, we first fine-tune each model for its target task using LoRA and then apply SOUL to the task-specific policies.
Quantitatively, SOUL increases clean success by an average of 20\%p while reducing hallucination failure by 22\%p across both settings.
As shown in Fig.~\ref{fig:fig_6}, the original policies exhibit state hallucination, accompanied by strong activation of forget features and weak activation of retain features.
After applying SOUL, policies complete the tasks without hallucination, with suppressed forget feature activations and increased retain feature activations.
These results show that SOUL remains effective in physical robot execution under diverse real-world visual conditions across both VLA architectures.

%
%
\input{Figures/Tex/7_tradeoff}
%
%

\subsection{Analyses}
{\textbf{Behavior preservation and correction.}}
To further examine the behavior-level changes underlying the reductions in hallucination and improvements in task success reported in TABLE~\ref{tab:tab_1}, we analyze the unlearning trade-off between preserving desirable behaviors and correcting unreliable behaviors.
Fig.~\ref{fig:fig_7} reports preservation and correction at both the task and event levels.
SOUL achieves an average preservation rate of 89\% across originally successful task executions and non-hallucinated events.
Meanwhile, it converts 40\% of originally failed task executions into successful ones and corrects 78\% of hallucinated events across the grasp, transport, and place stages.
These results demonstrate a favorable unlearning trade-off, substantially correcting undesirable hallucination behaviors while largely preserving behaviors that were already successful or hallucination-free.

%
%
\input{Tables/2_token}
%
%

{\textbf{Effect of region guidance.}}
To evaluate the effect of spatial grounding on sparse feature selection for policy unlearning, we compare global, tokenized, and region-guided feature selection using OpenVLA and $\pi_{0.5}$ on LIBERO-Plus and RoboCasa, respectively, with results averaged in TABLE~\ref{tab:tab_2}.
Global feature selection improves over the original policy, but is less effective than tokenized and region-guided selection, likely because pooling across visual tokens removes spatial information needed to isolate localized hallucination-associated patterns.
Tokenized selection performs better by preserving token-level localization, while region-guided selection achieves the lowest hallucination rates and the highest success rates.
These results show that preserving spatial structure during feature selection improves policy unlearning, with region guidance achieving the best balance between hallucination reduction and task success.

\textbf{Contributions of forget score and retain objective.}
To analyze the roles of the forget score \(s^{\mathcal F}_{r,j}\) and the retain objective \(\mathcal{L}_{\mathrm{retain}}\), we conduct ablations using OpenVLA and $\pi_{0.5}$ on LIBERO-Plus and RoboCasa, respectively, with results averaged in TABLE~\ref{tab:tab_3}.
The first row corresponds to the proposed method.
Without the retain objective (second row), hallucination reduction is maintained, but clean success decreases, showing its importance in preserving successful behavior.
Removing the aggregation score from the forget score (third row), both hallucination reduction and task success degrade, showing that incorporating the separation and success criteria enables more selective forgetting target selection.
Together, combining score aggregation with the retain objective achieves the best balance, supporting their complementary roles in selective policy unlearning.

%
%
\input{Tables/3_ablation}
%
%

\section{Conclusion}
We have presented a study of selective policy unlearning in VLAs, targeting state hallucination as a recurring undesirable execution behavior.
Our analysis showed that state hallucination is accompanied by weakened attention to task-relevant visual regions and distinct region-conditioned sparse feature patterns associated with hallucinated executions.
Based on these findings, we proposed \emph{SOUL}, which selectively unlearns policy knowledge associated with state hallucination by suppressing hallucination features while preserving success features.
Experiments across VLA architectures in simulated and real-world environments show that SOUL reduces state hallucination and improves task success while largely preserving existing manipulation capabilities.
These results show that interpretable sparse feature analysis can move beyond diagnosing VLA failures to provide actionable targets for policy unlearning, enabling selective modification of recurring undesirable behaviors in embodied robot policies.

\bibliographystyle{IEEEtran}
\bibliography{ref}

\end{document}

%% file: Figures/Tex/1_teaser.tex
\begin{figure}[t]
    \centering
    \captionsetup[subfigure]{font=small,justification=centering}
    \begin{subfigure}[t]{0.95\linewidth}
        \centering
        \includegraphics[width=\linewidth]{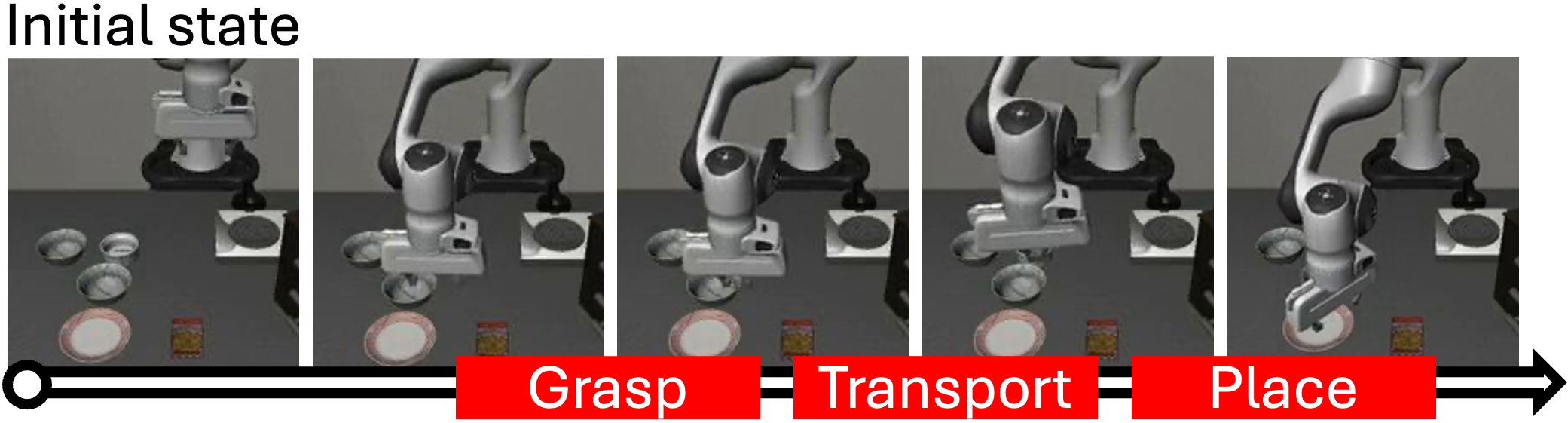}
        \caption{Representative example of state hallucination}
        \label{fig:1_teaser_ex}
    \end{subfigure}
    \\
    \begin{subfigure}[t]{0.65\linewidth}
        \centering
        \includegraphics[height=4.5cm]{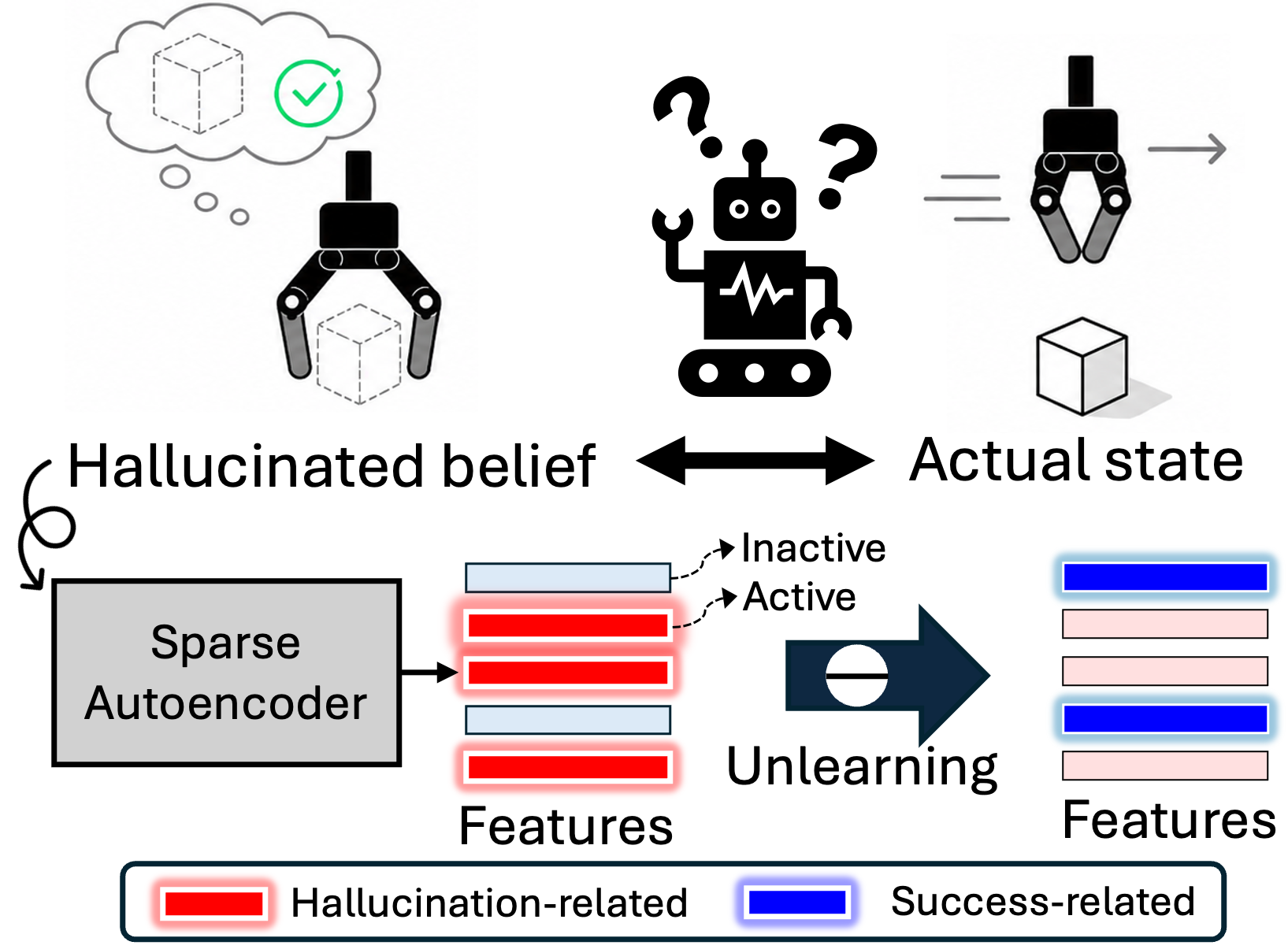}
        \caption{Overview of sparse feature policy unlearning for state hallucination}
        \label{fig:1_teaser_a}
    \end{subfigure}
    \hfill
    \begin{subfigure}[t]{0.25\linewidth}
        \centering
        \includegraphics[height=4.3cm]{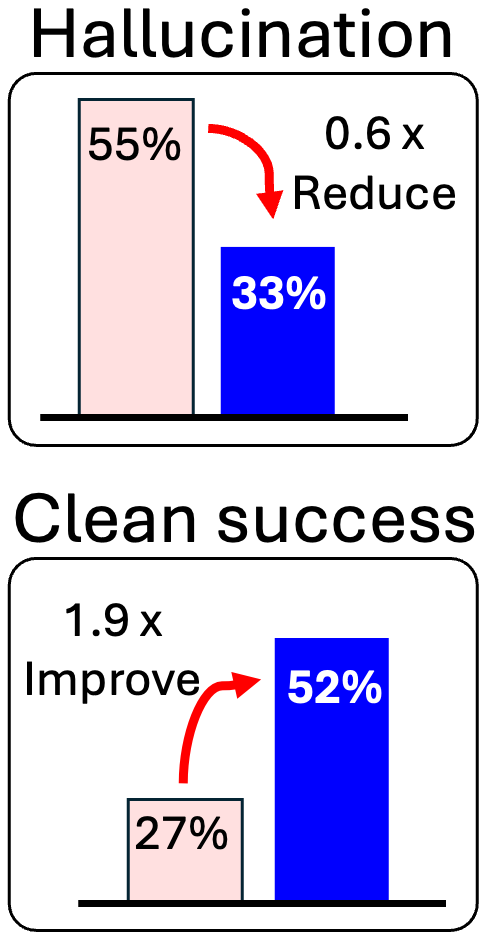}
        \caption{
        Performance after unlearning
        }
        \label{fig:1_teaser_b}
    \end{subfigure}
    \caption{\textbf{Sparse feature policy unlearning mitigates hallucinated beliefs in VLAs.}
    (a) Examples of grasp, transport, and place state hallucinations.
    (b) State hallucination occurs when a VLA mistakes its hallucinated belief for the actual robot-object state and generates actions based on this false belief.
    During this mismatch, hallucination features become strongly activated, and the proposed unlearning suppresses their activation while reinforcing success features.
    (c) Our sparse feature policy unlearning reduces hallucination while improving clean (hallucination-free) task success.
    }
    \label{fig:1_teaser}
\end{figure}

%% file: Figures/Tex/2_attention.tex
\begin{figure*}[t]
    \centering
    \begin{subfigure}{\textwidth}
        \centering
        \hspace{-3mm}%
        \begin{minipage}[t]{0.08\textwidth}
            \centering
            {\footnotesize\strut}\par
            \vspace{1mm}
    
            \parbox[c][1.7cm][c]{\linewidth}{
                \centering\footnotesize Clean success
            }\par
            \parbox[c][1.7cm][c]{\linewidth}{
                \centering\footnotesize Hallucination failure
            }
        \end{minipage}%
        \hspace{2mm}%
        %
        \begin{minipage}[t]{0.30\textwidth}
            \centering
            {\footnotesize\textbf{Gripper \& target object}}\par
            \vspace{1mm}
            \includegraphics[
                width=\linewidth,
                height=3.3cm
            ]{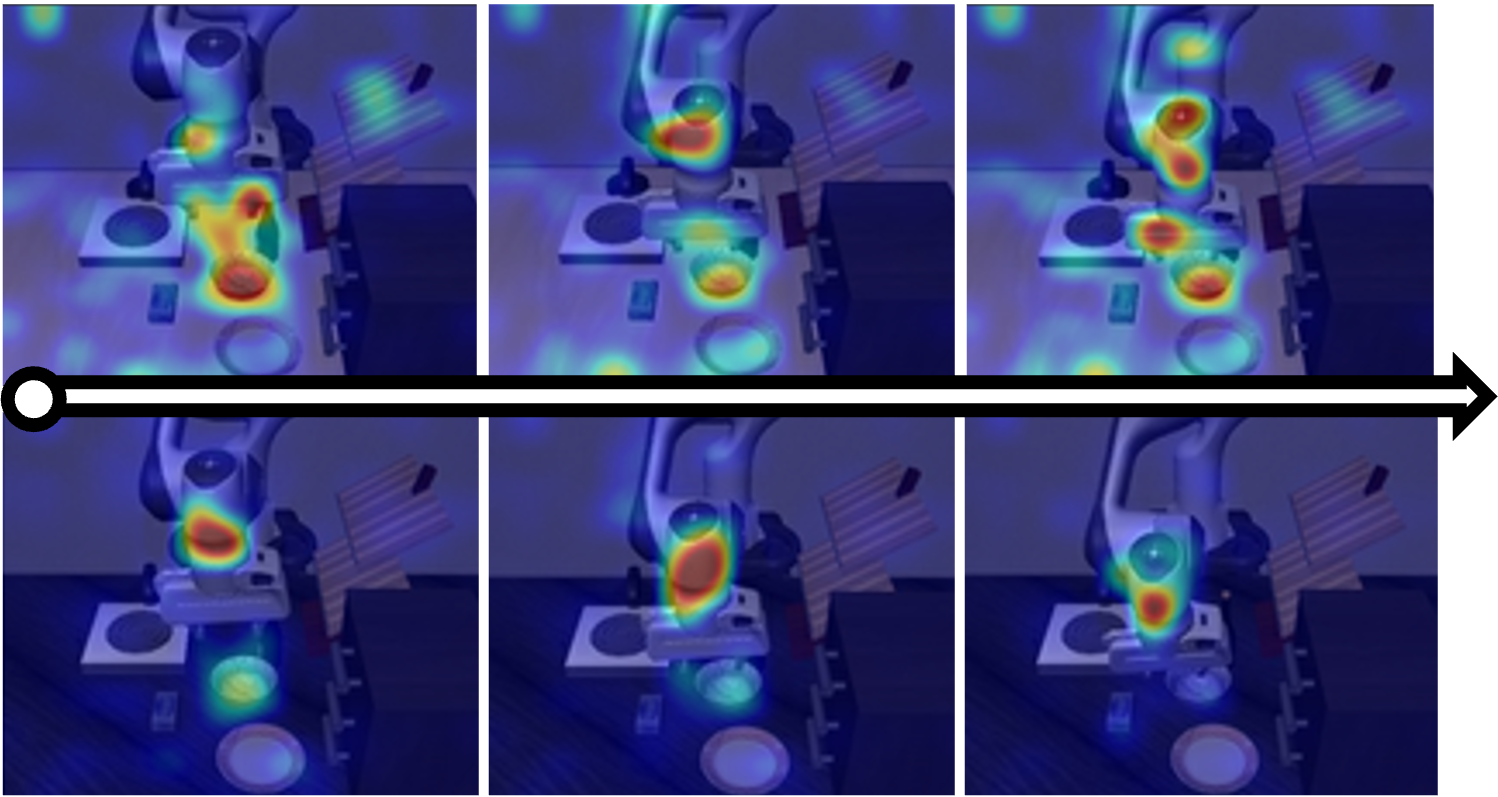}
        \end{minipage}%
        \hspace{1mm}%
        %
        \begin{minipage}[t]{0.30\textwidth}
            \centering
            {\footnotesize\textbf{Background \& target object}}\par
            \vspace{1mm}
            \includegraphics[
                width=\linewidth,
                height=3.3cm
            ]{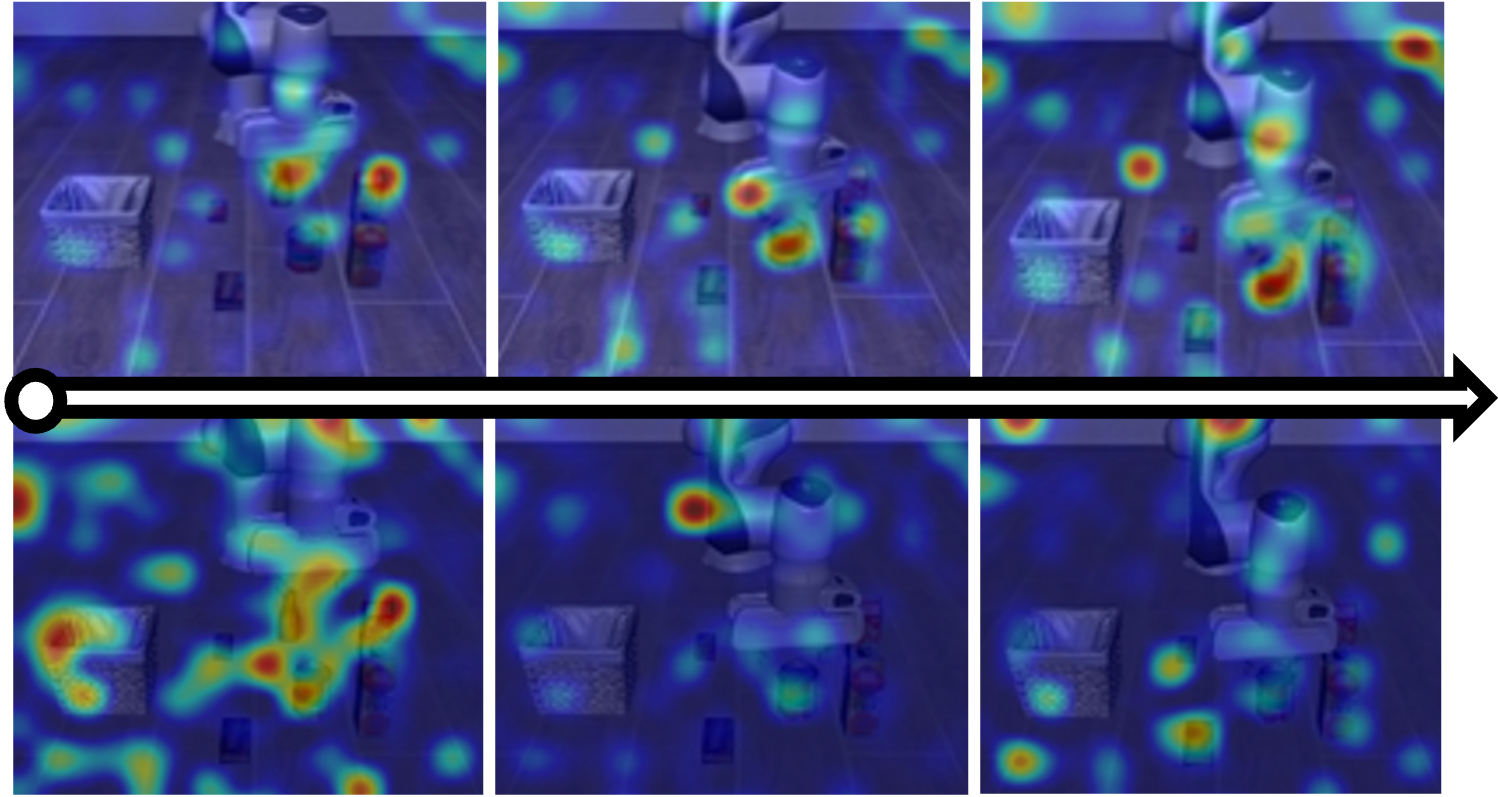}
        \end{minipage}%
        \hspace{1mm}%
        %
        \begin{minipage}[t]{0.30\textwidth}
            \centering
            {\footnotesize\textbf{Gripper \& Background}}\par
            \vspace{1mm}
            \includegraphics[
                width=\linewidth,
                height=3.3cm
            ]{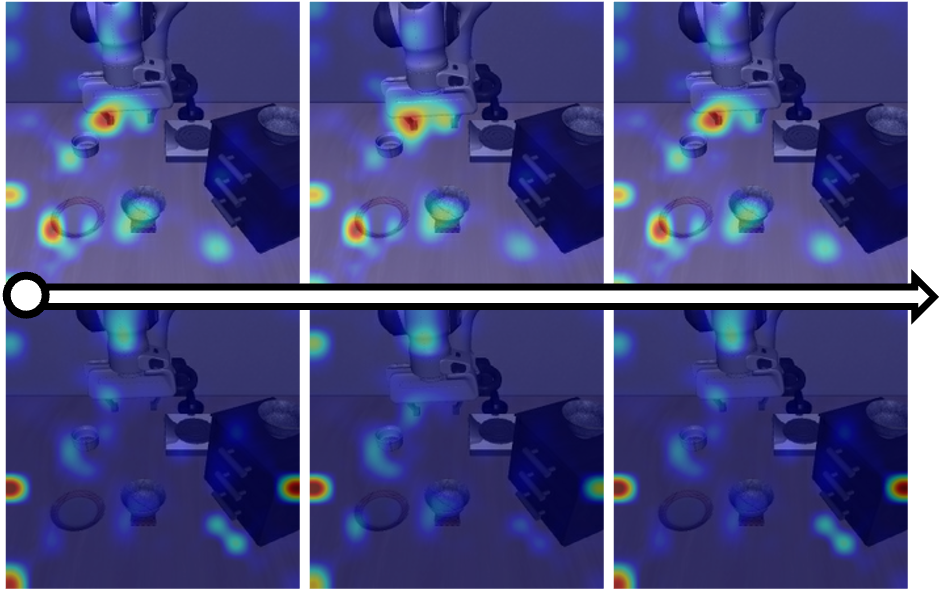}
        \end{minipage}
    \end{subfigure}
    \caption{
        \textbf{State hallucination is associated with distinct attention patterns.}
        Representative clean success (CS, hallucination-free success) and hallucination failure (HF, failure with hallucination) rollouts under the same task show distinct attention patterns.
        HF exhibits weaker attention to task-relevant regions (e.g., gripper and target object) and increased attention to the background.
        }
    \label{fig:2_attention}
\end{figure*}

%% file: Figures/Tex/3_feature.tex
\definecolor{csbg}{RGB}{228,235,247}
\definecolor{hfbg}{RGB}{248,233,233}
\definecolor{statbg}{RGB}{244,244,244}
\definecolor{cslabel}{RGB}{25,70,230}
\definecolor{hflabel}{RGB}{230,30,30}
\colorlet{cslegend}{csbg!70!cslabel}
\colorlet{hflegend}{hfbg!70!hflabel}

\newcommand{\featuretitlefont}{\footnotesize\bfseries}

\newlength{\featurepanelwidth}
\newlength{\featurestatwidth}
\newlength{\featurepairwidth}
\newlength{\featurelabelwidth}
\newlength{\featurecellgap}
\newlength{\featureimageheight}
\newlength{\featurestatshift}
\newsavebox{\featureimagebox}
\setlength{\featurelabelwidth}{20pt}
\setlength{\featurecellgap}{1.5pt}

\newcommand{\setfeatureimageheight}[1]{%
    \sbox{\featureimagebox}{\includegraphics[width=\featurepanelwidth]{#1}}%
    \setlength{\featureimageheight}{%
        \dimexpr\ht\featureimagebox+\dp\featureimagebox\relax}%
}

\newcommand{\featureheader}[3]{%
    \parbox[c][0.34cm][c]{\linewidth}{%
        \noindent
        \makebox[\featurelabelwidth][l]{\hspace{1.5pt}#1}%
        \makebox[\dimexpr\linewidth-2\featurelabelwidth\relax][c]{%
            \adjustbox{max width=\dimexpr\linewidth-2\featurelabelwidth\relax}{#2}%
        }%
        \makebox[\featurelabelwidth][r]{#3\hspace{1.5pt}}%
    }%
}

\newcommand{\paneltitleline}[4]{%
    \featureheader
        {\textcolor{#1}{\scriptsize\bfseries #2}}
        {#4 #3}
        {}%
}

\newcommand{\featurepanel}[6]{%
    \cellcolor{#1}%
    \featurepanelbody{#2}{#3}{#4}{#5}{#6}%
}

\newcommand{\featurepanelbody}[5]{%
    \begin{minipage}[t]{\linewidth}
        \centering
        \vspace{0pt}
        \paneltitleline{#1}{#2}{#3}{#4}\par
        \vspace{0.5pt}
        \includegraphics[width=\linewidth]{#5}\par
        \vspace{0pt}
    \end{minipage}%
}

\newcommand{\statcell}[1]{%
    \cellcolor{statbg}%
    \begin{minipage}[t]{\linewidth}
        \centering
        \vspace{0pt}
        \parbox[c][0.34cm][c]{\linewidth}{%
            \centering\footnotesize\bfseries Statistics
        }\par
        \vspace{0.5pt}
        \makebox[\linewidth][c]{%
            \hspace*{-\featurestatshift}%
            \includegraphics[width=0.94\linewidth,
                height=\featureimageheight,keepaspectratio]{#1}%
            \hspace*{\featurestatshift}%
        }\par
        \vspace{0pt}
    \end{minipage}%
}

\newcommand{\successhfpanel}[2]{%
    \cellcolor{hfbg}%
    \begin{minipage}[t]{\linewidth}
        \centering
        \vspace{0pt}
        \parbox[c][0.34cm][c]{\linewidth}{%
            \noindent
            \makebox[\linewidth][r]{%
                \makebox[0pt][r]{%
                    \makebox[\featurepairwidth][c]{%
                        \adjustbox{max width=\dimexpr\featurepairwidth-2\featurelabelwidth\relax}{%
                            \featuretitlefont #1%
                        }%
                    }%
                }%
            }%
            \makebox[0pt][r]{%
                \textcolor{hflabel}{\scriptsize\bfseries HF}\hspace{1.5pt}%
            }%
        }\par
        \vspace{0.5pt}
        \includegraphics[width=\linewidth]{#2}\par
        \vspace{0pt}
    \end{minipage}%
}

\newcommand{\legendchip}[1]{%
    \begingroup
    \setlength{\fboxsep}{0pt}%
    \colorbox{#1}{\rule{0pt}{1.2ex}\rule{5ex}{0pt}}%
    \endgroup
}

\newcommand{\featurelegend}{%
    \begingroup
    \setlength{\fboxsep}{2pt}%
    \fcolorbox{black!35}{white}{%
        \adjustbox{max width=\dimexpr\linewidth-2\fboxsep-2\fboxrule\relax}{%
            \footnotesize
            \renewcommand{\arraystretch}{1}%
            \begin{tabular}{@{}l@{\hspace{10pt}}l@{\hspace{10pt}}l@{\hspace{10pt}}l@{}}
                \legendchip{cslegend}\hspace{3pt}\mbox{CS rollout} &
                \legendchip{hflegend}\hspace{3pt}\mbox{HF rollout} &
                \legendchip{blue}\hspace{3pt}\mbox{Success event} &
                \legendchip{red}\hspace{3pt}\mbox{Hallucination event}
            \end{tabular}%
        }%
    }%
    \endgroup
}

\begin{figure*}[t]
    \centering
    \setlength{\tabcolsep}{0pt}
    \renewcommand{\arraystretch}{0}
    \setlength{\extrarowheight}{0pt}
    \setlength{\featurestatwidth}{0.15\textwidth}
    \setlength{\featurestatshift}{0.10\featurestatwidth}
    \setlength{\featurepanelwidth}{%
        \dimexpr\textwidth/2-\featurestatwidth/2-\featurecellgap\relax}
    \setlength{\featurepairwidth}{%
        \dimexpr2\featurepanelwidth+\featurecellgap\relax}

    \noindent\makebox[\textwidth][l]{%
        \begin{minipage}{\featurepairwidth}
            \centering
            \featurelegend
        \end{minipage}%
    }\par
    \vspace{2pt}

    \begin{subfigure}{\textwidth}
        \centering
        \setfeatureimageheight{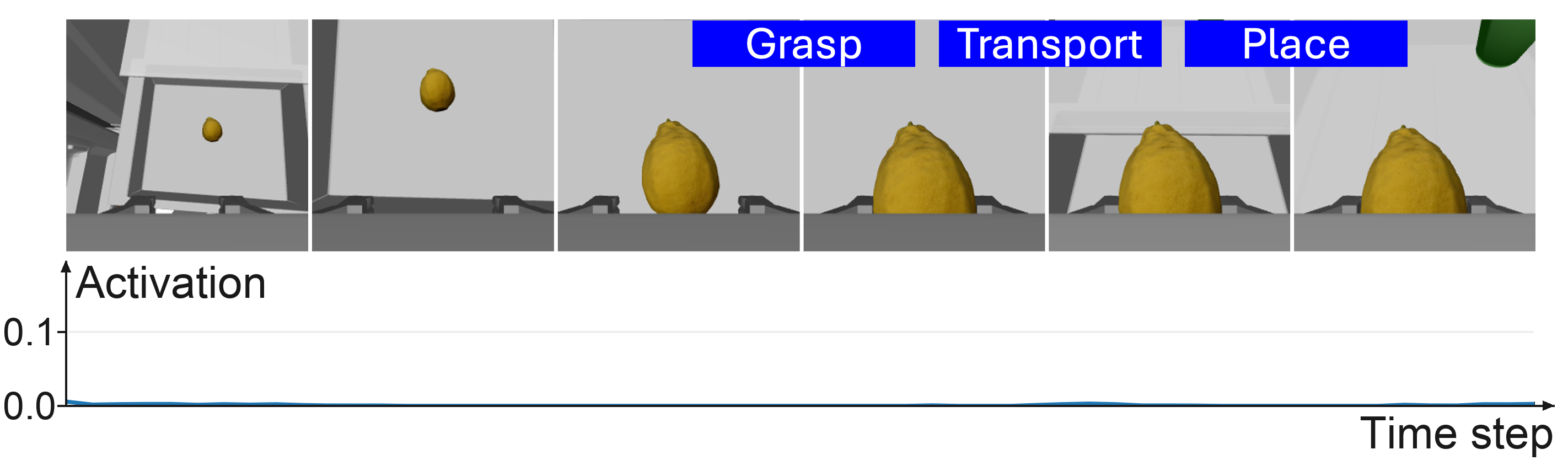}
        \begin{tabular}{@{}
            >{\centering\arraybackslash}p{\featurepanelwidth}
            @{\hspace{\featurecellgap}}
            >{\centering\arraybackslash}p{\featurepanelwidth}
            @{\hspace{\featurecellgap}}
            >{\centering\arraybackslash}p{\featurestatwidth}
            @{}}
            \featurepanel
                {csbg}{cslabel}{CS}
                {}{\featuretitlefont}
                {Figures/3_hall_cs.png}
            &
            \successhfpanel
                {PickPlaceFridgeDrawerToShelf}
                {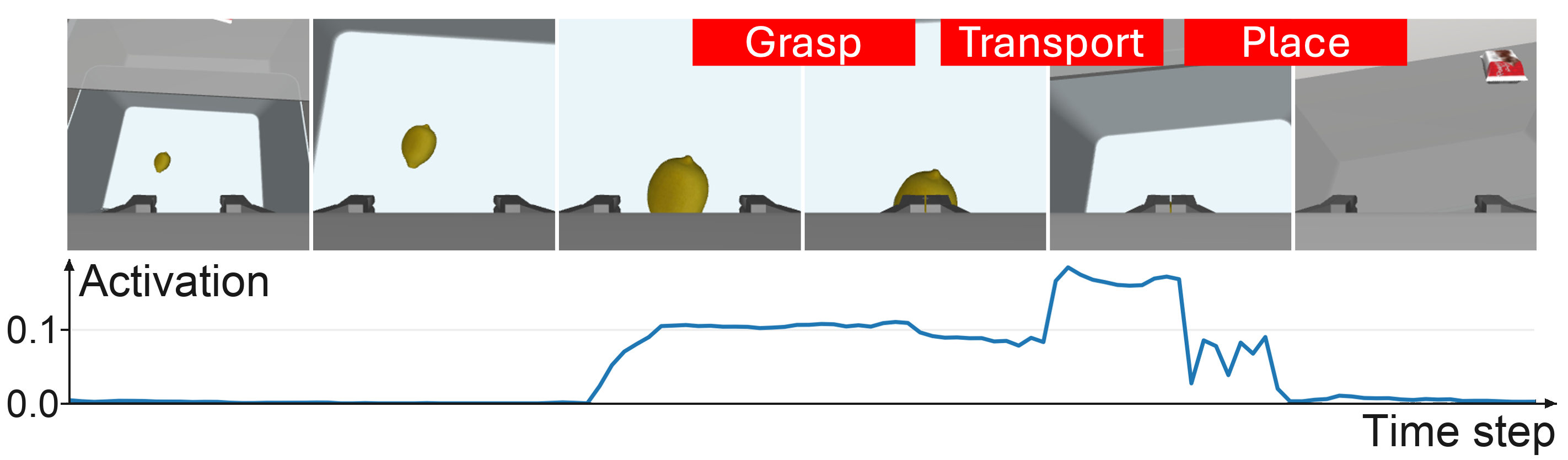}
            &
            \statcell{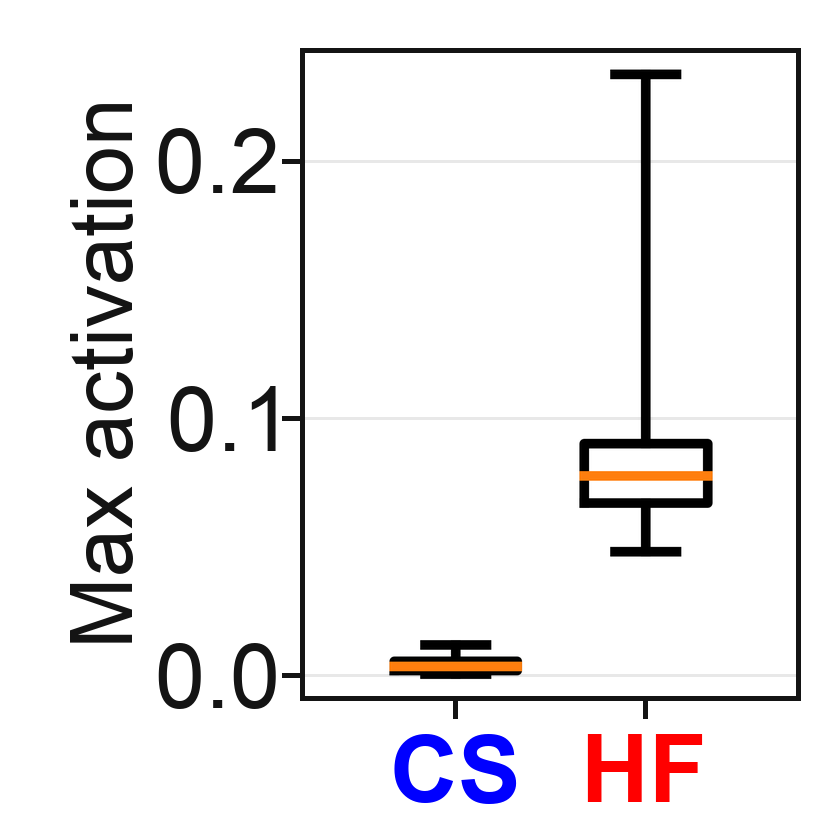}
            \\
            \noalign{\vskip\featurecellgap}
            \featurepanel
                {hfbg}{hflabel}{HF}
                {PickPlaceDrawerToCounter}
                {\featuretitlefont}
                {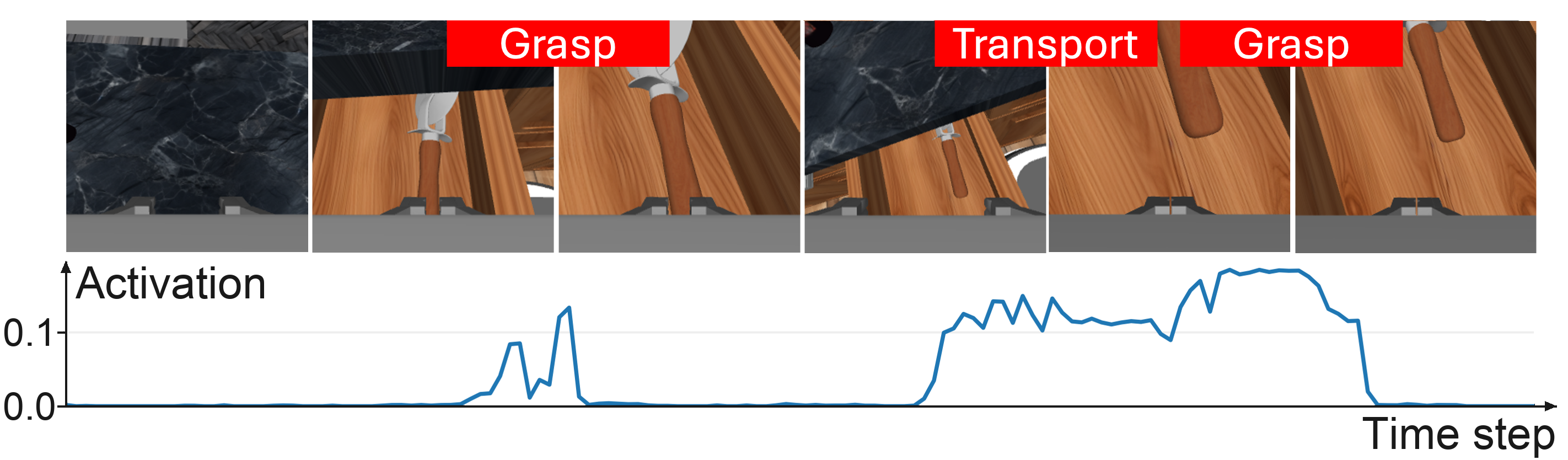}
            &
            \begingroup
            \renewcommand{\paneltitleline}[4]{%
                \featureheader
                    {}%
                    {#4 #3}%
                    {\textcolor{#1}{\scriptsize\bfseries #2}}%
            }
            \featurepanel
                {hfbg}{hflabel}{HF}
                {PickPlaceCounterToSink}
                {\featuretitlefont}
                {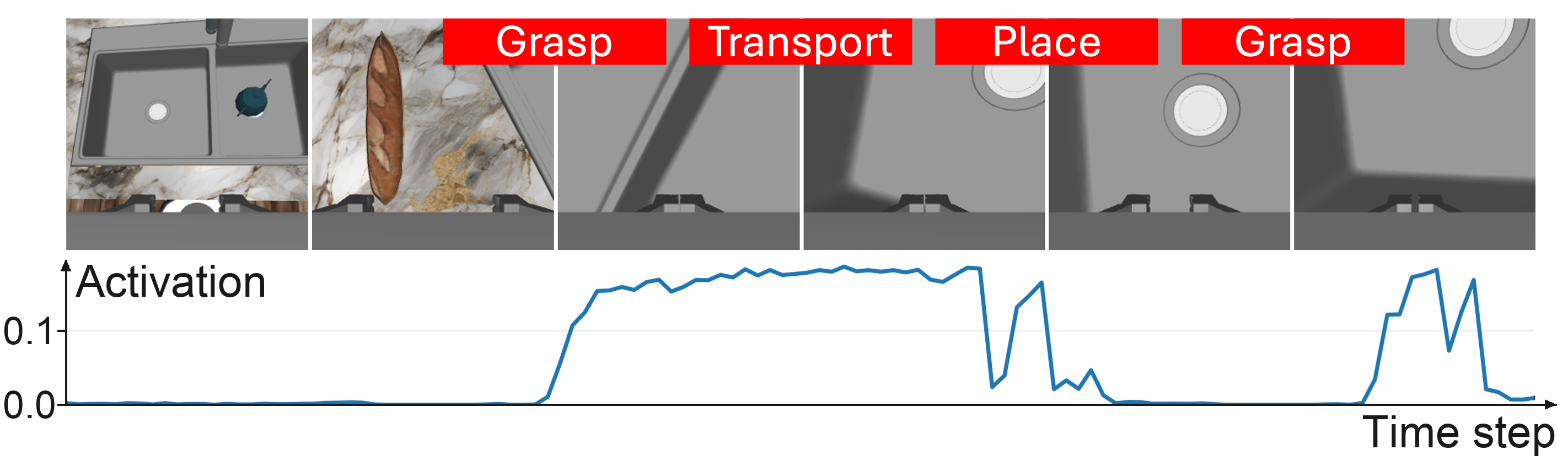}
            \endgroup
            &
            {}
        \end{tabular}
        \caption{Hallucination feature}
    \end{subfigure}

    \vspace{2mm}

    \begin{subfigure}{\textwidth}
        \centering
        \setfeatureimageheight{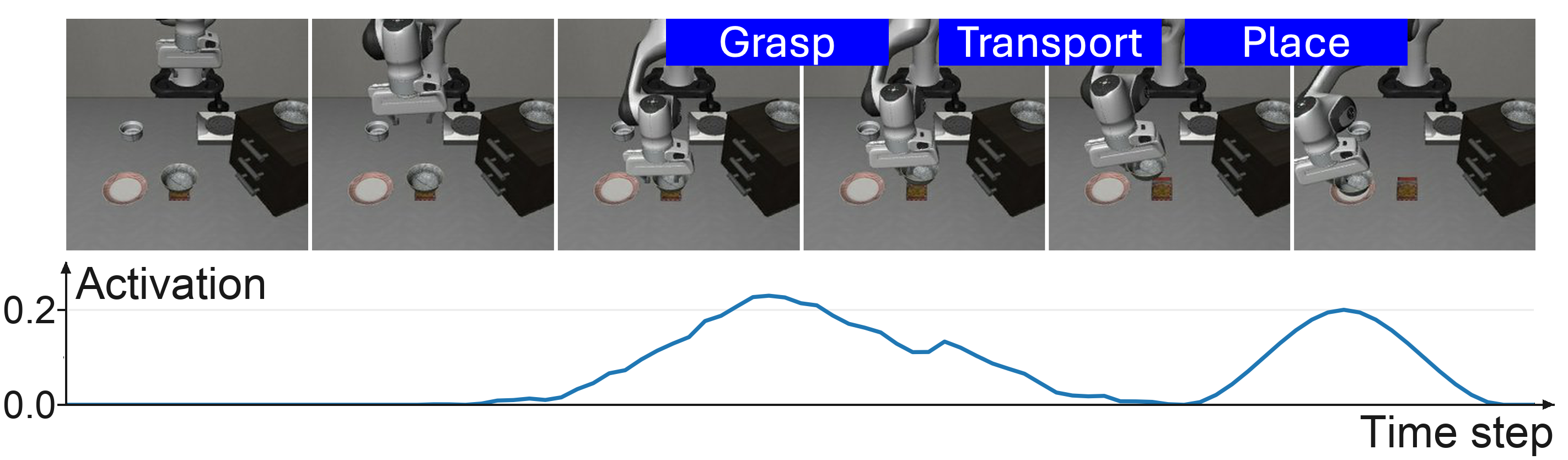}
        \begin{tabular}{@{}
            >{\centering\arraybackslash}p{\featurepanelwidth}
            @{\hspace{\featurecellgap}}
            >{\centering\arraybackslash}p{\featurepanelwidth}
            @{\hspace{\featurecellgap}}
            >{\centering\arraybackslash}p{\featurestatwidth}
            @{}}
            \featurepanel
                {csbg}{cslabel}{CS}
                {}{\featuretitlefont}
                {Figures/3_clean_cs.png}
            &
            \successhfpanel
                {Pick up the black bowl on the cookie box and place it on the plate.}
                {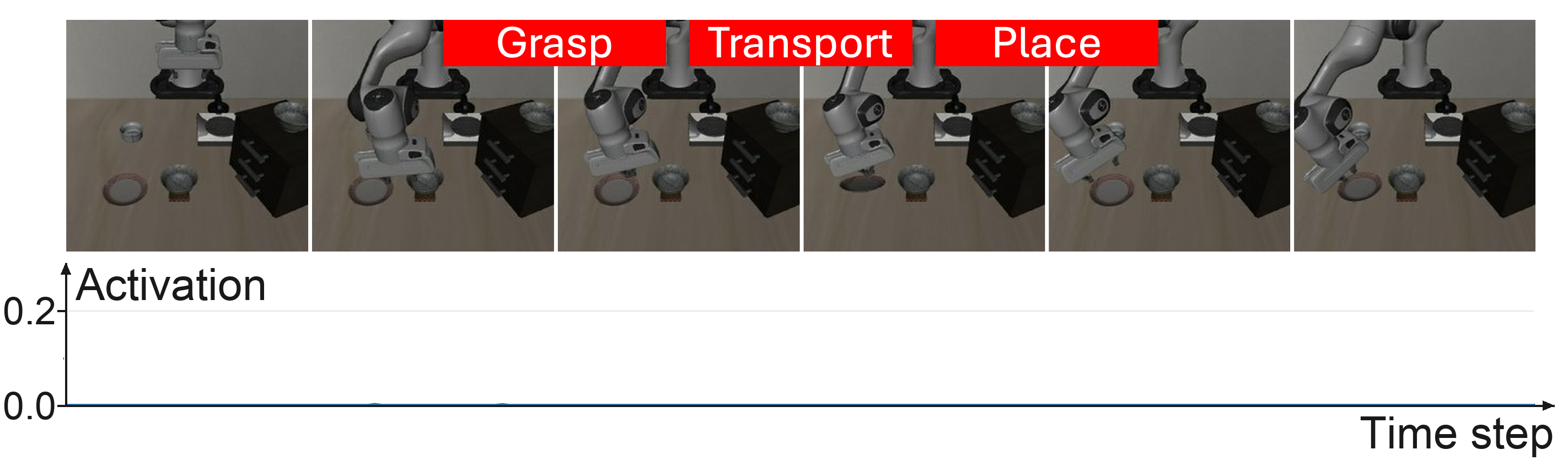}
            &
            \statcell{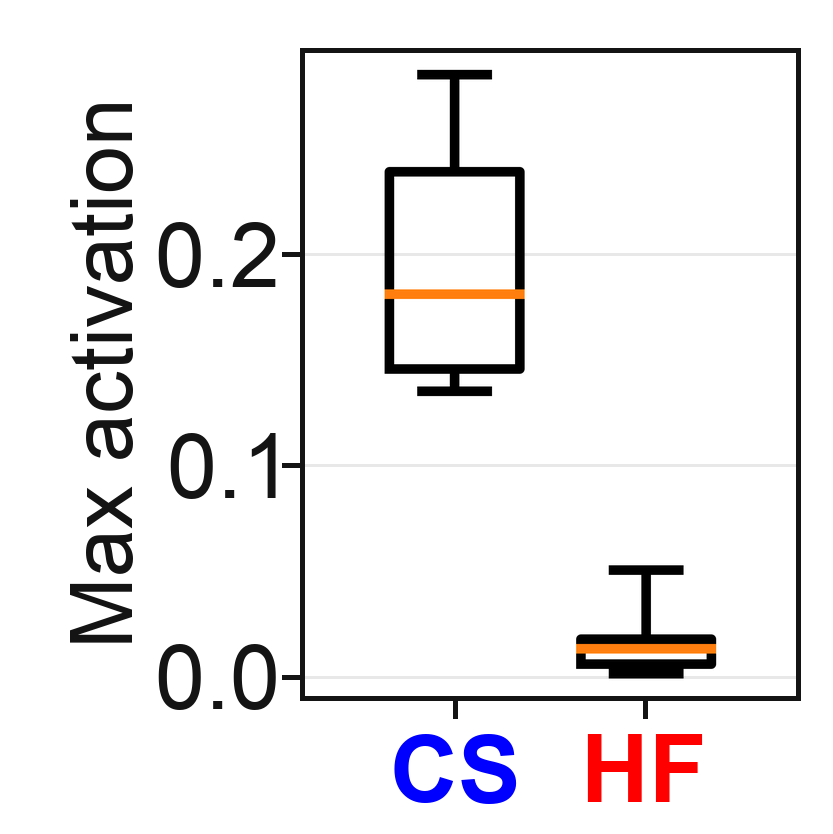}
        \end{tabular}
        \caption{Success feature}
    \end{subfigure}
    \caption{%
    \textbf{Sparse features associated with state hallucination can be identified.}
    For each feature, we show clean success (CS) and hallucination failure (HF) rollouts with their temporal activations, \(z_{e,t,r,j}\), together with the distribution of rollout-level maximum activations, \(m_{e,r,j}\).
    (a) A hallucination feature becomes strongly activated around hallucination events, while remaining weak in CS.
    (b) A success feature is strongly activated during successful subtask execution or task completion in CS, while remaining weak in HF.
    The boxplots show consistent activation differences between CS and HF across rollouts.
    }
    \label{fig:3_feature}
\end{figure*}

%% file: Figures/Tex/4_overview.tex
\begin{figure*}[t]
    \centering
    \includegraphics[width=0.9\textwidth]{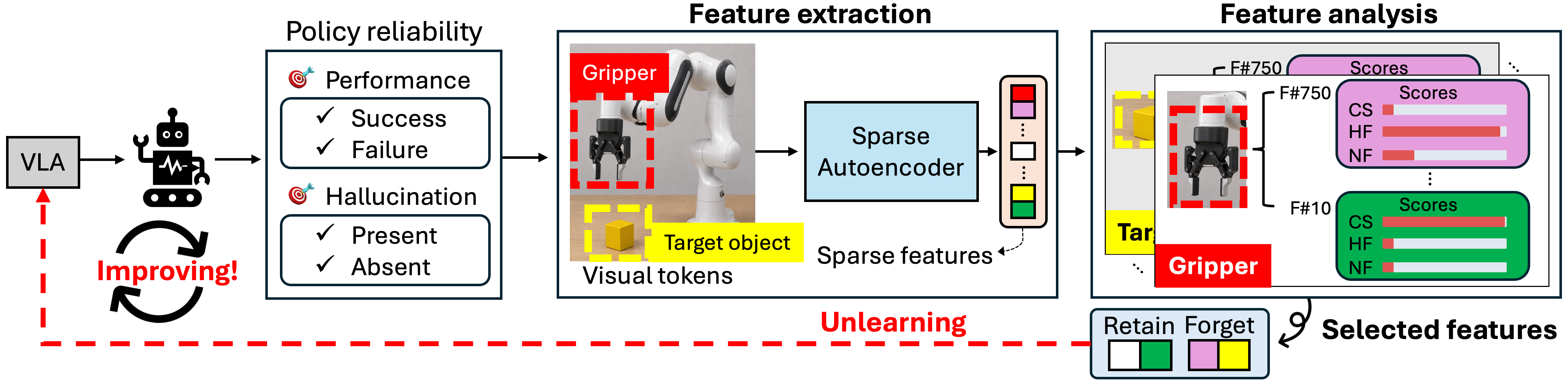}
    \caption{
    \textbf{Overall framework.}
    The proposed method provides a framework for improving robot policies by identifying and selectively modifying internal representations associated with undesirable behaviors.
    It first extracts visual-region-specific sparse features from policy activations and analyzes their activation patterns across execution outcomes, including clean success (CS), hallucination failure (HF), and non-hallucination failure (NF), characterized by task performance and hallucination occurrence.
    Based on these associations, region-feature pairs are selected as forget or retain targets.
    The policy is then updated to suppress hallucination-associated features while preserving reliable behavior.
    In this way, the policy is improved.
}    
\label{fig:fig_4_overview}
\end{figure*}

%% file: Tables/1_main.tex
\begin{table*}[t]
\centering
\caption{
\textbf{Performance comparison of hallucination and task success on benchmarks.}
All values are reported as percentages (\%).
`Hall.' denotes hallucination.
$\downarrow$ and $\uparrow$ indicate that lower and higher values are better, respectively.
}
\label{tab:tab_1}
\resizebox{\linewidth}{!}{%
\begin{tabular}{c||c|c|c|c||c|c|c}
\hline
Method
& \textbf{Hallucination $\downarrow$} 
& \makecell[c]{\textbf{Hallucination} \\ \textbf{failure} $\downarrow$} 
& \makecell[c]{\textbf{Overall} \\ \textbf{success} $\uparrow$} 
& \makecell[c]{\textbf{Clean} \\ \textbf{success} $\uparrow$}
& \makecell[c]{\textbf{Grasp}\\ Hall. $\downarrow$} 
& \makecell[c]{\textbf{Transport}\\ Hall. $\downarrow$} 
& \makecell[c]{\textbf{Place}\\ Hall. $\downarrow$} \\
\hline
\rowcolor{gray!15}
\multicolumn{8}{c}{{LIBERO-Plus with OpenVLA}} \\
\hline
Original Policy~\cite{kim2025openvla}
& 70.0 & 60.0 & 26.0 & 16.0 & 70.0 & 38.0 & 16.0 \\
Gradient Ascent~\cite{maini2024tofu}
& 84.0 & 84.0 & 0.0 & 0.0 & 84.0 & 62.0 & 0.0 \\
\textbf{SOUL (Ours)}
&\textbf{44.0} & \textbf{38.0} & \textbf{58.0} & \textbf{52.0} & \textbf{44.0} & \textbf{20.0} & 6.0 \\
\hline
\rowcolor{gray!15}
\multicolumn{8}{c}{{RoboCasa with $\pi_{0.5}$}} \\
\hline
Original Policy~\cite{intelligence2025pi_}
& 40.0 & 37.1 & 41.9 & 39.0 & 39.0 & 7.6 & 1.9 \\
Gradient Ascent~\cite{maini2024tofu}
& 60.0 & 59.0 & 4.8 & 3.8 & 60.0 & 0.0 & 0.0 \\
\textbf{SOUL (Ours)}
& \textbf{22.9} & \textbf{21.0} & \textbf{54.3} & \textbf{52.4} & \textbf{22.9} & 3.8 & \textbf{0.0} \\
\hline
\end{tabular}
}
\end{table*}

%% file: Figures/Tex/5_benchmark_rollouts.tex
\begin{figure*}[t]
    \centering
    \captionsetup[subfigure]{font=small,justification=centering}

    \newlength{\qualH}
    \setlength{\qualH}{2cm}

    \begin{minipage}[c]{0.07\textwidth}
        \centering
    \end{minipage}
    \hspace{0.01\textwidth}
    \begin{minipage}[c]{0.45\textwidth}
        \centering
        \hspace{2.2cm}\footnotesize\textbf{Before unlearning (Original policy)}
    \end{minipage}
    \hfill
    \begin{minipage}[c]{0.45\textwidth}
        \centering
         \footnotesize\textbf{After unlearning (SOUL, proposed method)}
    \end{minipage}

    \vspace{1mm}

    \begin{subfigure}{\textwidth}
        \centering
        \hspace{-3mm}
    
        \begin{minipage}[c][\qualH][c]{0.07\textwidth}
            \centering
            \footnotesize
            Retain\\[1.3cm]
            Forget
        \end{minipage}
        %
        \begin{minipage}[c]{0.44\textwidth}
            \centering
            \includegraphics[width=\linewidth,height=\qualH]{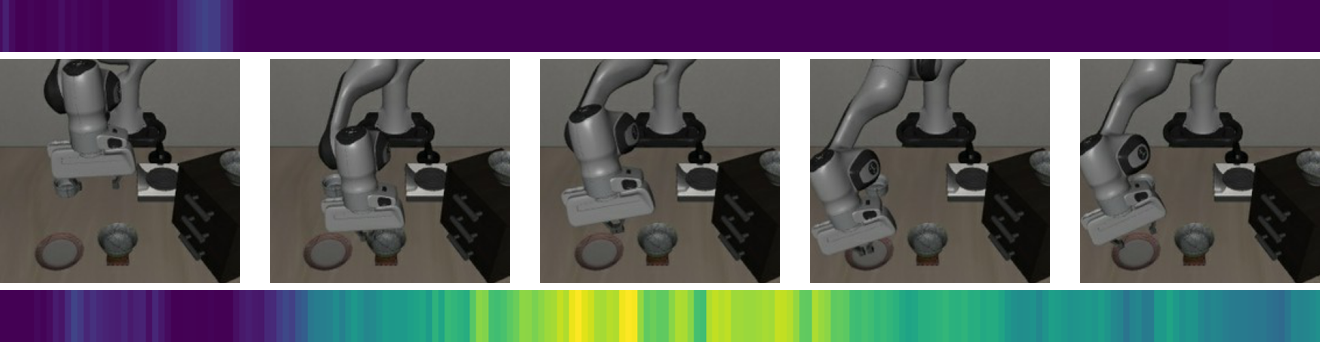}
        \end{minipage}
        \begin{tikzpicture}[baseline=-0.5ex]
            \draw[->, line width=1.5pt] (0,0) -- (0.25,0);
        \end{tikzpicture}
        %
        \begin{minipage}[c]{0.44\textwidth}
            \centering
            \includegraphics[width=\linewidth,height=\qualH]{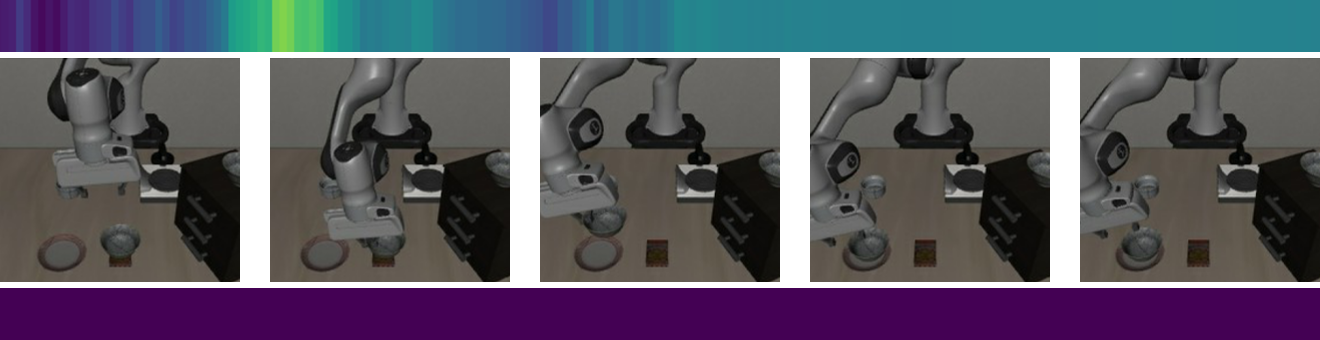}
        \end{minipage}
        \caption{LIBERO-Plus with OpenVLA}
        \label{fig:fig_5_libero}
    \end{subfigure}
    \begin{subfigure}{\textwidth}
        \centering
        \hspace{-3mm}
        \begin{minipage}[c][\qualH][c]{0.07\textwidth}
            \centering
            \footnotesize
            Retain\\[1.3cm]
            Forget
        \end{minipage}
        %
        \begin{minipage}[c]{0.44\textwidth}
            \centering
            \includegraphics[width=\linewidth,height=\qualH]{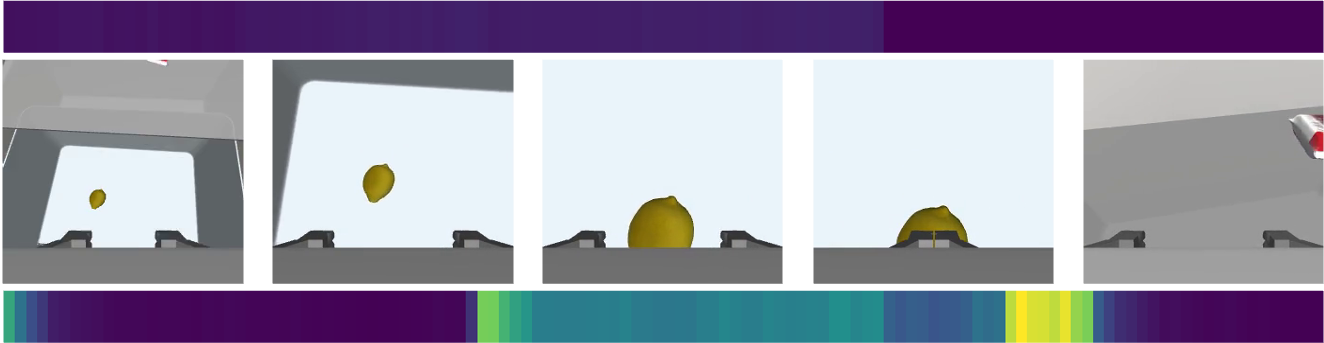}
        \end{minipage}
        \begin{tikzpicture}[baseline=-0.5ex]
            \draw[->, line width=1.5pt] (0,0) -- (0.25,0);
        \end{tikzpicture}
        %
        \begin{minipage}[c]{0.44\textwidth}
            \centering
            \includegraphics[width=\linewidth,height=\qualH]{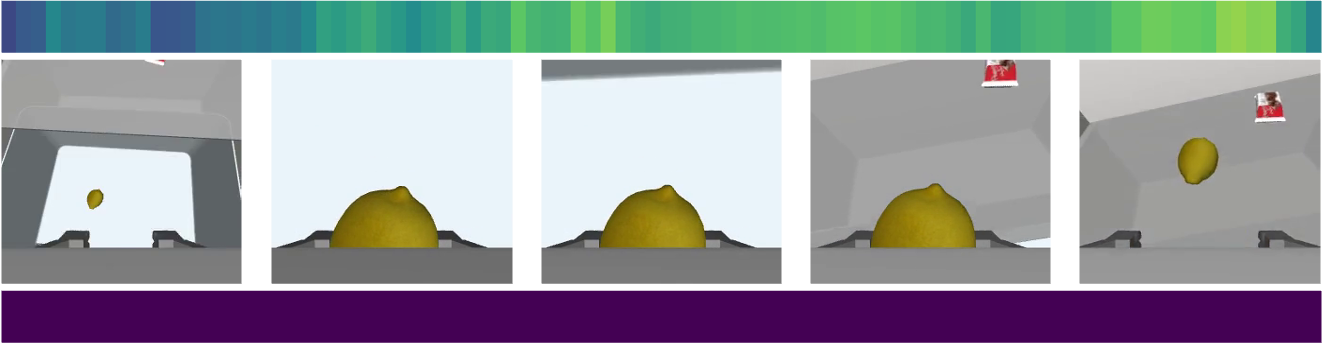}
        \end{minipage}

    \caption{
        RoboCasa with $\pi_{0.5}$
        \makebox[0pt][l]{%
            \hspace{0.28\textwidth}%
            {\scriptsize Activation:}\,
            \raisebox{-0.8mm}{
                \includegraphics[width=0.8cm,height=0.35cm]{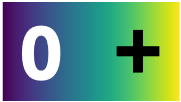}
            }
        }
    }
    \label{fig:fig_5_robocasa}
    \end{subfigure}
    \caption{
    \textbf{Visualizations of rollout behavior and feature activations on benchmarks.}
    We compare the original policy and the proposed method, showing the temporal activations of selected forget and retain features during execution.
    }
    \label{fig:fig_5}
\end{figure*}

%% file: Figures/Tex/6_real_rollouts.tex
\begin{figure*}[t]
    \centering
    \captionsetup[subfigure]{font=small,justification=centering}

    \newlength{\realH}
    \setlength{\realH}{2cm}

    \begin{minipage}[c]{0.07\textwidth}
        \centering
    \end{minipage}
    \hspace{0.01\textwidth}
    \begin{minipage}[c]{0.44\textwidth}
        \centering
        \hspace{2.2cm}\footnotesize\textbf{Before unlearning (Original policy)}
    \end{minipage}
    \hfill
    \begin{minipage}[c]{0.44\textwidth}
        \centering
        \footnotesize\textbf{After unlearning (SOUL, proposed method)}
    \end{minipage}

    \vspace{1mm}

    \begin{subfigure}{\textwidth}
        \centering
        \hspace{-3mm}

        \begin{minipage}[c][\realH][c]{0.07\textwidth}
            \centering
            \footnotesize
            Retain\\[1.3cm]
            Forget
        \end{minipage}
        %
        \begin{minipage}[c]{0.44\textwidth}
            \centering
            \includegraphics[
                width=\linewidth,
                height=\realH
            ]{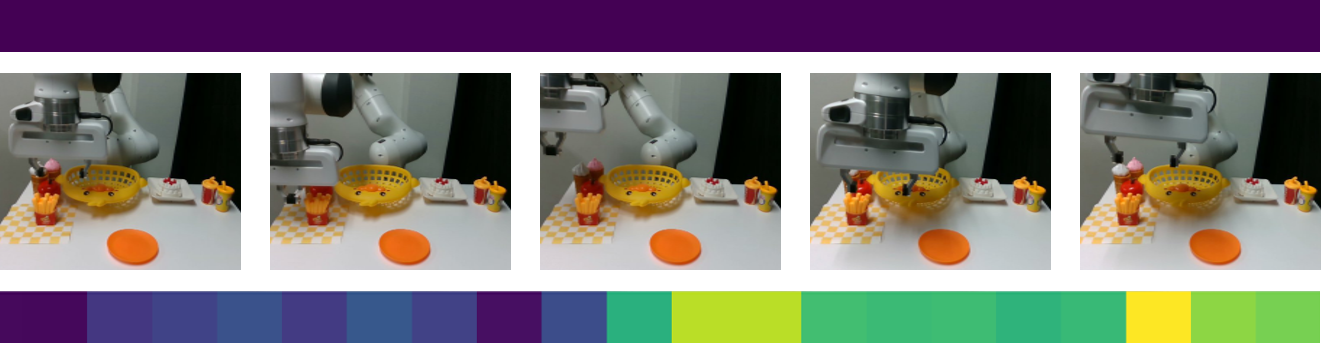}
        \end{minipage}
        \begin{tikzpicture}[baseline=-0.5ex]
            \draw[->, line width=1.5pt] (0,0) -- (0.25,0);
        \end{tikzpicture}
        %
        %
        \begin{minipage}[c]{0.44\textwidth}
            \centering
            \includegraphics[
                width=\linewidth,
                height=\realH
            ]{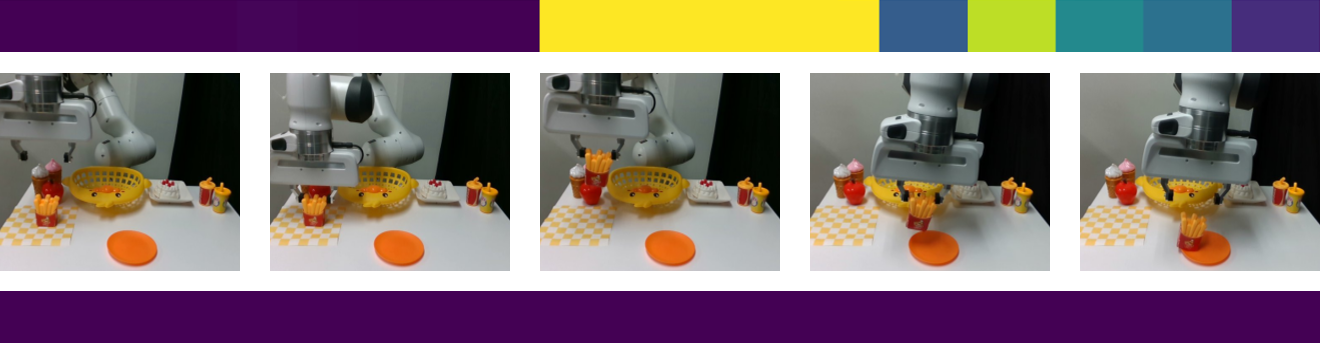}
        \end{minipage}
        \caption{\emph{`Pick fries and put it on the plate'} with OpenVLA}

        \label{fig:fig_6_rollout1}
    \end{subfigure}
    \begin{subfigure}{\textwidth}
        \centering
        \hspace{-3mm}

        \begin{minipage}[c][\realH][c]{0.07\textwidth}
            \centering
            \footnotesize
            Retain\\[1.3cm]
            Forget
        \end{minipage}
        %
        \begin{minipage}[c]{0.44\textwidth}
            \centering
            \includegraphics[
                width=\linewidth,
                height=\realH
            ]{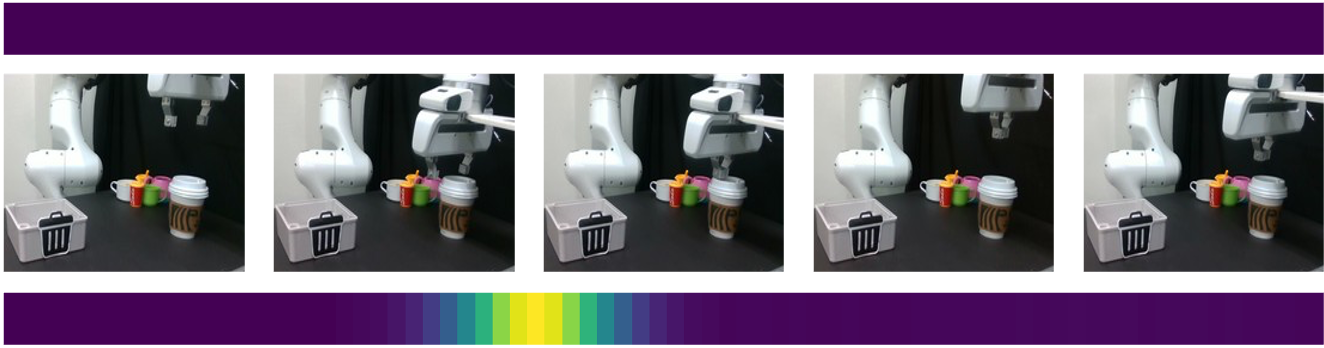}
        \end{minipage}
        \begin{tikzpicture}[baseline=-0.5ex]
            \draw[->, line width=1.5pt] (0,0) -- (0.25,0);
        \end{tikzpicture}
        %
        %
        \begin{minipage}[c]{0.44\textwidth}
            \centering
            \includegraphics[
                width=\linewidth,
                height=\realH
            ]{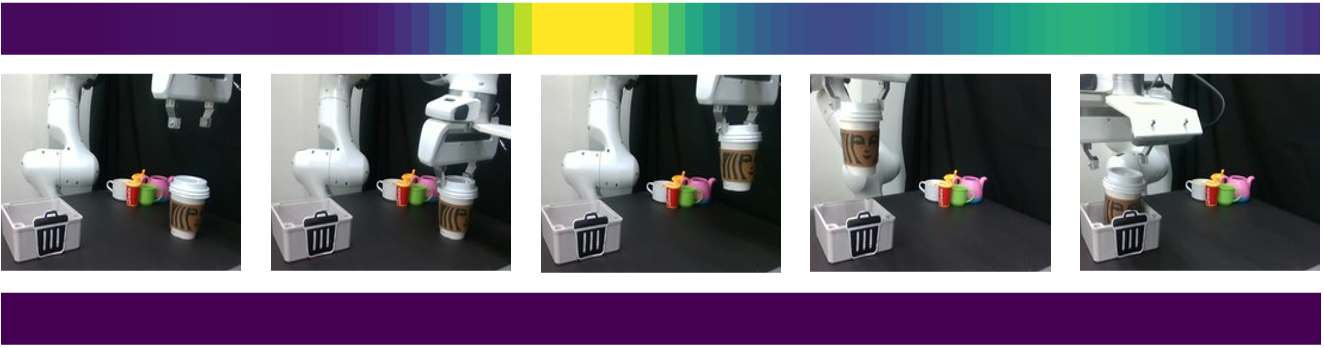}
        \end{minipage}

    \caption{
        \emph{`Throw the coffee in the trash bin'} with $\pi_{0.5}$
        \makebox[0pt][l]{%
            \hspace{0.19\textwidth}%
            {\scriptsize Activation:}\,
            \raisebox{-0.8mm}{
                \includegraphics[width=0.8cm,height=0.35cm]{Figures/activation_legend.png}
            }
        }
    }

        \label{fig:fig_6_rollout2}
    \end{subfigure}
    \caption{
    \textbf{Real-world rollouts and selected sparse feature activations before and after applying the proposed unlearning, SOUL.}
    We compare the original policy and SOUL across representative real-world rollouts spanning both models and tasks.
    }
    \label{fig:fig_6}
\end{figure*}

%% file: Figures/Tex/7_tradeoff.tex
\begin{figure}[t]
    \centering

    \includegraphics[
        width=0.9\columnwidth
    ]{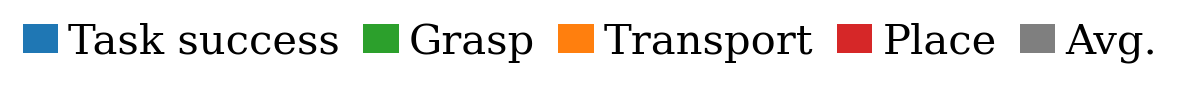}
    \begin{subfigure}[t]{0.48\columnwidth}
        \centering
        \includegraphics[
            width=\linewidth,
        ]{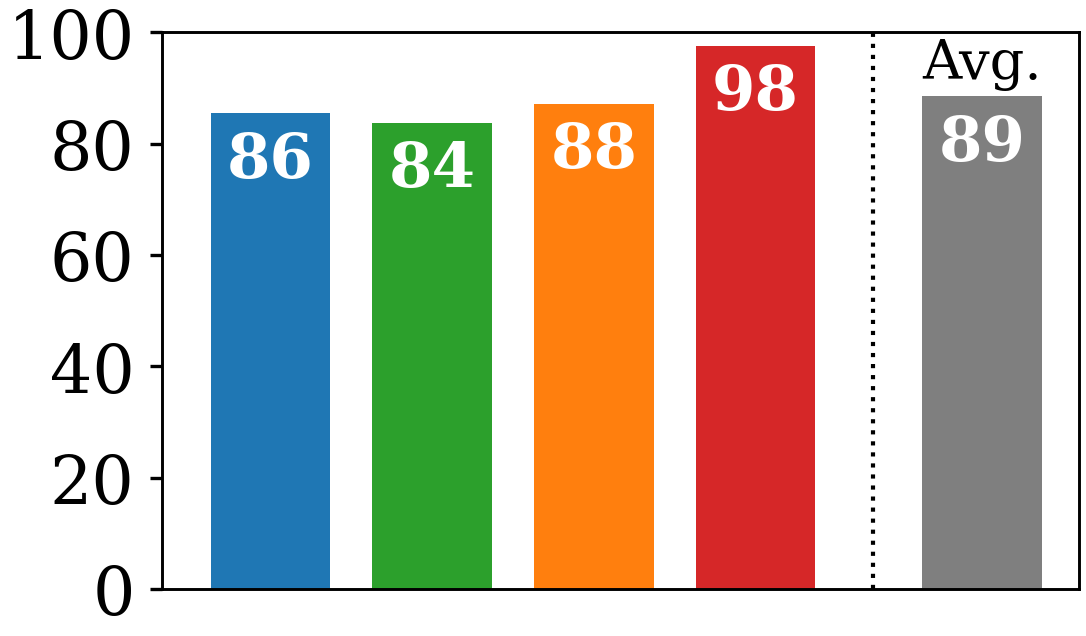}
        \caption{Preservation $\uparrow$}
        \label{fig:fig_7_preserve}
    \end{subfigure}
    \begin{subfigure}[t]{0.48\columnwidth}
        \centering
        \includegraphics[
            width=\linewidth,
        ]{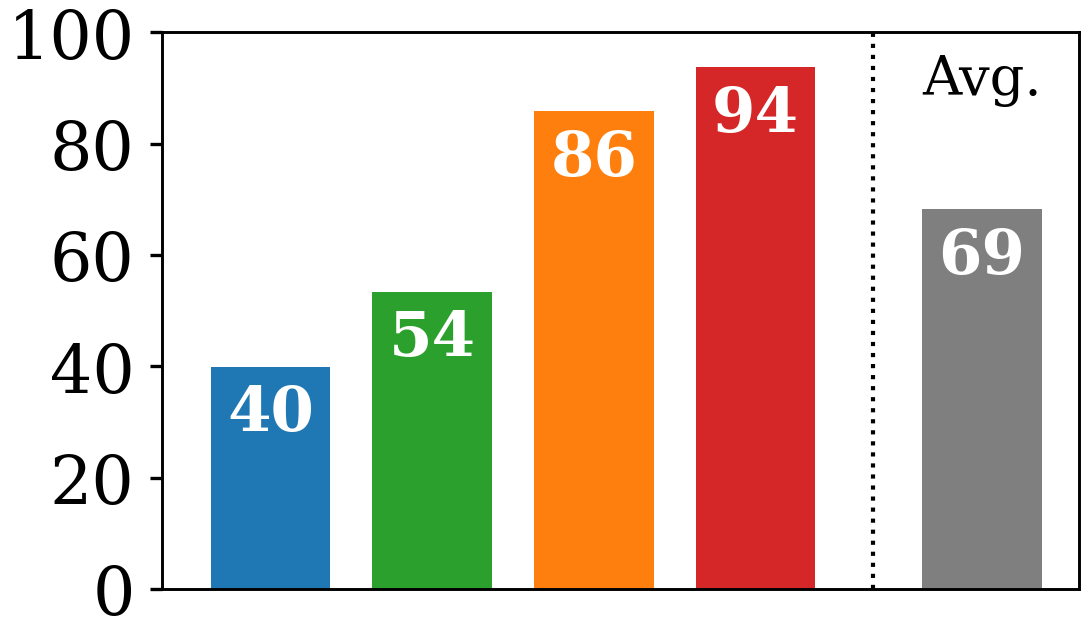}
        \caption{Correction $\uparrow$}
        \label{fig:fig_7_improvement}
    \end{subfigure}
    \caption{
    \textbf{Behavior preservation and correction after applying SOUL.}
    (a) Preservation measures the fraction of originally successful task executions and non-hallucinated events that remain successful and non-hallucinated, respectively, after SOUL.
    (b) Correction measures the fraction of originally failed task executions and hallucinated events that become successful and non-hallucinated, respectively, after SOUL.
    Higher values indicate better preservation or correction.
    }
    \label{fig:fig_7}
\end{figure}

%% file: Tables/2_token.tex
\begin{table}[t]
\centering
\caption{
\textbf{Comparison of feature selection at different spatial granularities.}
`Global' selects features from activations pooled across all visual tokens, `Tokenized' from individual visual-token activations, and `Region-guided' from activations aggregated within each semantic region.
}
\label{tab:tab_2}

\setlength{\tabcolsep}{4pt}
\renewcommand{\arraystretch}{0.90}

\resizebox{\columnwidth}{!}{%
\begin{tabular}{c||cccc}
\hline
Method 
& \textbf{Hall. $\downarrow$} 
& \makecell[c]{\textbf{Hall.} \\ \textbf{failure} $\downarrow$} 
& \makecell[c]{\textbf{Overall} \\ \textbf{success} $\uparrow$} 
& \makecell[c]{\textbf{Clean} \\ \textbf{success} $\uparrow$}
\\
\hline

Global
& 45.8
& 41.4
& 42.7
& 38.3
\\

Tokenized
& 41.5
& 35.5
& 47.3
& 41.3
\\

Region-guided
& \textbf{33.5}
& \textbf{29.5}
& \textbf{56.2}
& \textbf{52.2}
\\

\hline
\end{tabular}%
}
\end{table}

%% file: Tables/3_ablation.tex
\begin{table}[t]
\centering
\caption{
\textbf{Ablation study of the forget score formulation and retain objective.}
The aggregated score \(s^{agg}_{r,j}=s^{sep}_{r,j}-s^{succ}_{r,j}\) complements \(s^{hall}_{r,j}\) for forget feature selection.
}
\label{tab:tab_3}

\setlength{\tabcolsep}{3pt}
\renewcommand{\arraystretch}{0.90}

\resizebox{\columnwidth}{!}{%
\begin{tabular}{ccc||cccc}
\hline
$s^{\mathrm{hall}}_{r,j}$ 
& $s^{\mathrm{agg}}_{r,j}$ 
& $\mathcal{L}_{\mathrm{retain}}$
& \textbf{Hall. $\downarrow$} 
& \makecell[c]{\textbf{Hall.} \\ \textbf{failure} $\downarrow$} 
& \makecell[c]{\textbf{Overall} \\ \textbf{success} $\uparrow$} 
& \makecell[c]{\textbf{Clean} \\ \textbf{success} $\uparrow$}
\\
\hline

\checkmark & \checkmark & \checkmark
& \textbf{33.5}
& \textbf{29.5}
& \textbf{56.2}
& \textbf{52.2}
\\

\checkmark & \checkmark &
& 36.8
& 29.9
& 55.2
& 48.3
\\

\checkmark & & \checkmark
& 39.9
& 36.9
& 50.7
& 47.7
\\

\hline
\end{tabular}%
}
\end{table}